\documentclass[12pt]{article}
\usepackage[top=1in, bottom=1in, left=1in, right=1in]{geometry}                
\usepackage{graphicx}
\usepackage{bm,amssymb,amsmath,amsthm}
\usepackage{epstopdf}
\usepackage{verbatim}
\usepackage{multirow}
\usepackage{setspace}
\usepackage{threeparttable}
\usepackage{textcomp}
\usepackage{tikz}
\usetikzlibrary{positioning}
\usepackage{hyperref}
\usepackage{color}
\usepackage{bbm}
\usepackage[natbibapa]{apacite}

\newtheorem{procedure}{Procedure}

\title{An Exploratory Analysis of the Latent Structure of Process Data via Action Sequence Autoencoders}
\author{Xueying Tang, Zhi Wang, Jingchen Liu, and Zhiliang Ying}
\date{}

\begin{document}
\maketitle

\begin{abstract}
Computer simulations have become a popular tool of assessing complex skills such as problem-solving skills. Log files of computer-based items record the entire human-computer interactive processes for each respondent. The response processes are very diverse, noisy, and of nonstandard formats. Few generic methods have been developed for exploiting the information contained in process data.
In this article, we propose a method to extract latent variables from process data. The method utilizes a sequence-to-sequence autoencoder to compress response processes into standard numerical vectors. It does not require prior knowledge of the specific items and human-computers interaction patterns. The proposed method is applied to both simulated and real process data to demonstrate that the resulting latent variables extract useful information from the response processes.
\end{abstract}

\section{Introduction}\label{sec:intro}

Problem solving is one of the key skills for people in the current world full of rapid changes \citep{OECD2017problem}. Computer-based items have recently become popular for assessing problem solving skills. In such items, problem-solving scenarios can be conveniently simulated through human-computer interfaces and the problem-solving processes can be easily recorded for analysis.




\begin{figure}[htb]
\includegraphics[width=\textwidth]{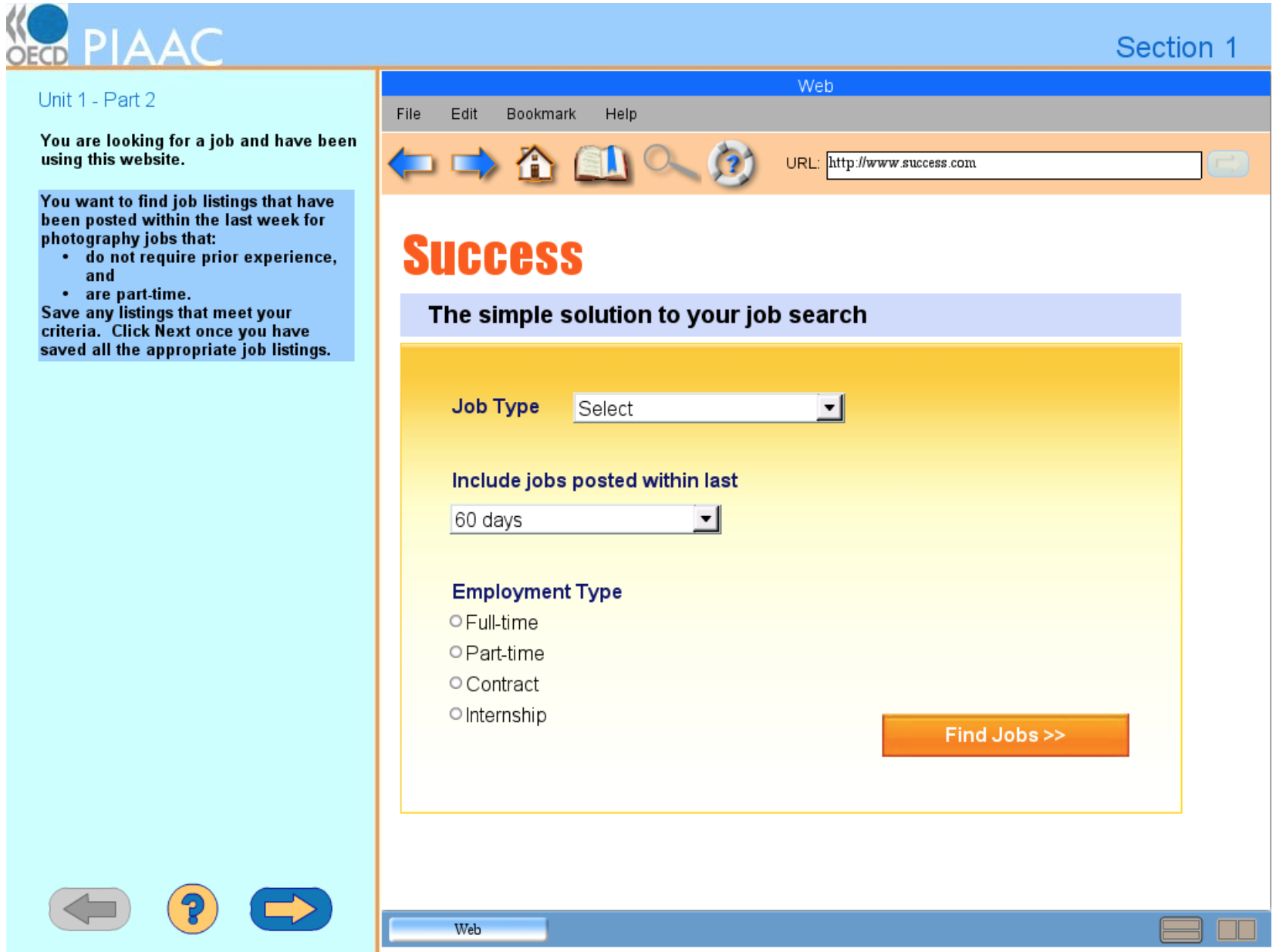}
\caption{Main page of the sample item.}\label{fig:item_main}
\end{figure}

\begin{figure}[htb]
\includegraphics[width=\textwidth]{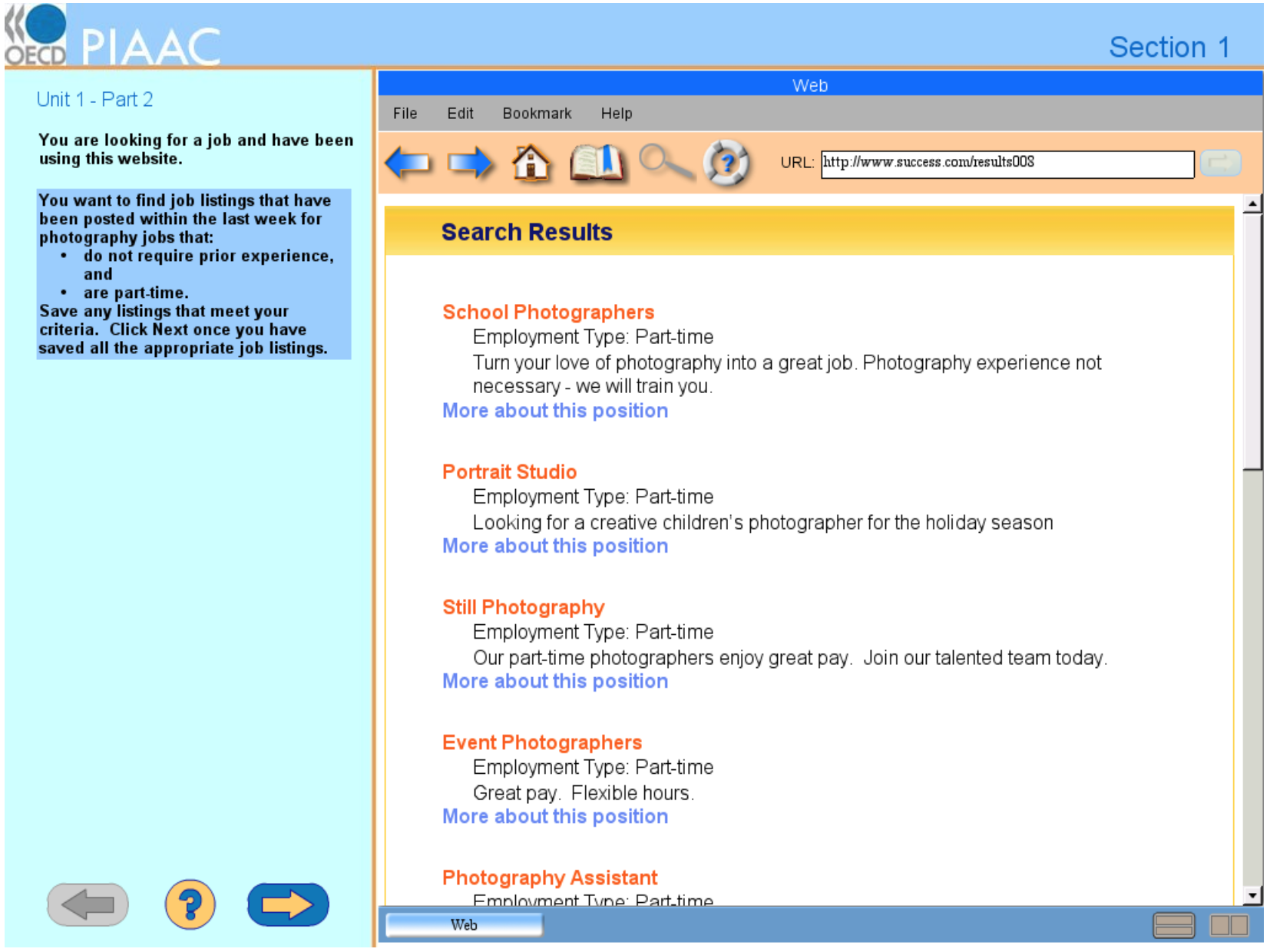}
\caption{Webpage after clicking ``Find Jobs'' in Figure \ref{fig:item_main}.}\label{fig:item_page1}
\end{figure}

\begin{figure}[htb]
\includegraphics[width=\textwidth]{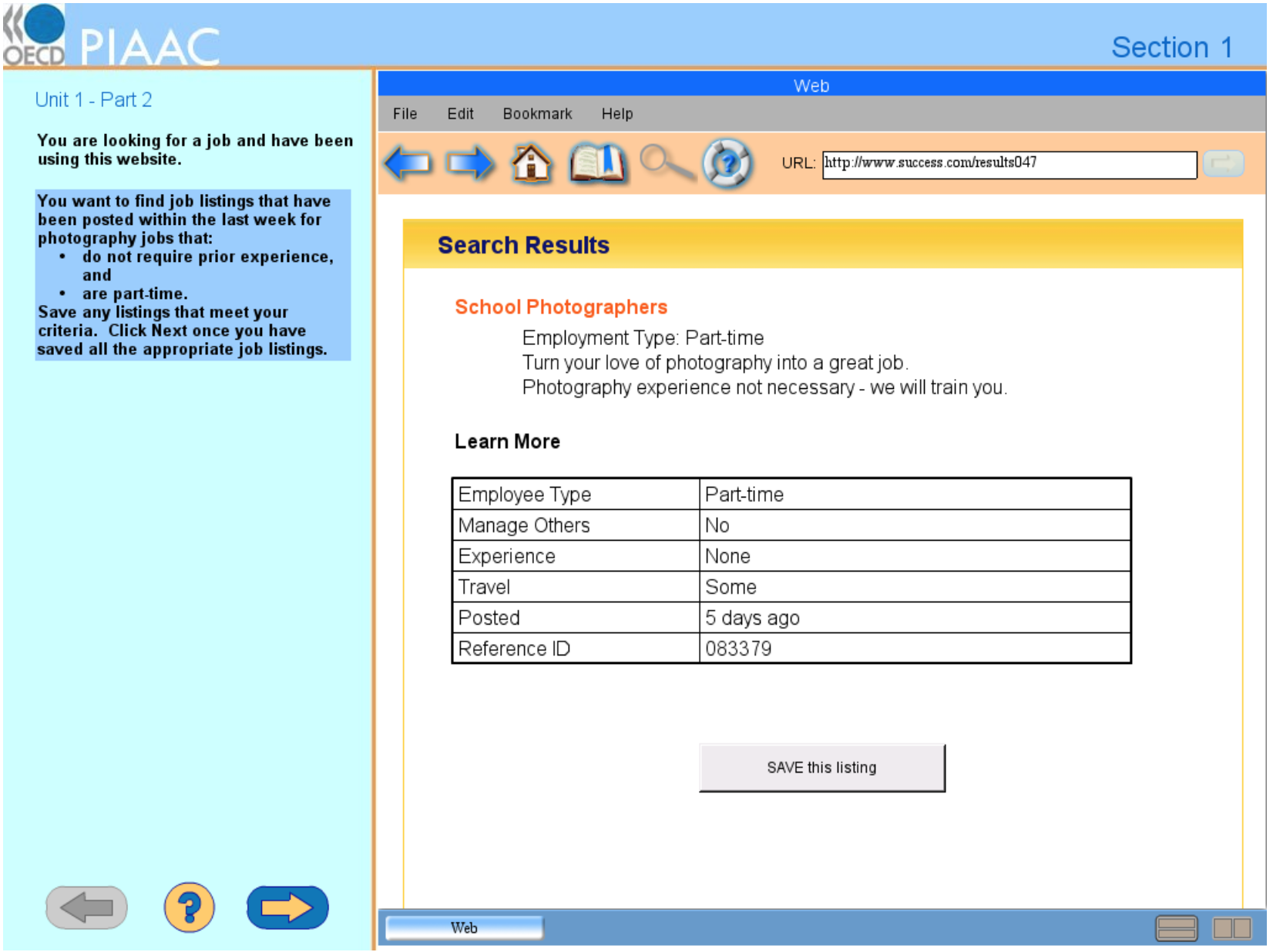}
\caption{Detailed information page of the first job listing in Figure \ref{fig:item_page1}.}\label{fig:item_page2}
\end{figure}

In 2012, several computer-based items were designed and deployed in the Programme for International Assessment of Adult Competencies (PIAAC) to measure adults' competency in problem solving in technology-rich environments (PSTRE). Screenshots\footnote{Retrieved from \url{https://piaac-logdata.tba-hosting.de/}} of the interface of a released PSTRE item are shown in Figures \ref{fig:item_main}--\ref{fig:item_page2}. The opening page of the item, displayed in Figure \ref{fig:item_main}, consists of two panels. The left panel contains item instructions and navigation buttons, while the right panel is the main medium for interaction. In this example, the right panel is a web browser showing a job searching website. The task is to find all job listings that meet the criteria described in the instructions. The dropdown menus and radio buttons can be used to narrow down the search range. Once the ``Find Jobs'' button is clicked, jobs that meet the selected criteria will be listed on the web page as shown in Figure \ref{fig:item_page1}. Participants can read the detail information about a listing by clicking ``More about this position''. Figure \ref{fig:item_page2} is the detailed information page of the first listing in Figure \ref{fig:item_page1}. If a listing is considered to meet all the requirements, it can be saved by clicking the ``SAVE this listing'' button. When a participant works on a problem, the entire response process is recorded in the log files in additional to the final response outcome (correct/incorrect). For example, if a participant selected ``Photography'' and ``7 days'' in the two dropdown menus, clicked the ``Part-time'' radio button, then clicked ``Find Jobs'', read the detailed information of the first listing and saved it, then a sequence of actions, ``Start, Dropdown1\_Photography, Dropdown2\_7, Part-time, Find\_Jobs, Click\_W1, Save, Next'', is recorded in the log files\footnote{This example item is not used in real practice. The coding of the actions and the action sequence described above were created for illustration purpose.}. The entire action sequence constitutes a single observation of process data. It tracks all the major actions the participants took when they interacted with the browsing environment.

%

The process responses contain substantially more comprehensive information of respondents than the traditional item responses that is often dichotomous (correct/incorrect) or polytomous (partial credits). 
On the other hand, to what extent this information is useful for educational and cognitive assessments and how to systematically make full use of such information are largely unknown. 
One of the difficulties in analyzing process data is to cope with its nonstandard format. Each process is a sequence of categorical variables (mouse clicks and keystrokes) and its length varies across observations. As a result, existing models for traditional item responses such as item response theory (IRT) models are not directly applicable to process data. Although some models have been extended to incorporate item response time \citep{entink2009multivariate,wang2018using,zhan2018cognitive}, similar extensions for response processes are difficult.

Another challenge for analyzing process data comes from the wide diversity of human behaviors. Signals from behavioral patterns in response processes are often attenuated by a large amount of noisy actions. The 1-step or 2-step lagged correlation of the response processes are often close to zero, indicating that models only capturing short-term dependence are often inadequate. 

The rich variety of computer-based items also adds to the difficulty in developing general methods for process data. The computer interface involved in the PSTRE items in PIAAC 2012 includes web browser, mail clients, and spreadsheet. The required tasks in these items also vary greatly. 
In some recent development of process data analysis such as \citet{greiff2016understanding} and \citet{kroehne2018conceptualize}, process data are first summarized into several variables according to domain knowledge and then their relationship with other variables of interest are investigated by conventional statistical methods. The design of the summary variables is usually item-specific and requires a thorough understanding of respondents' cognitive process during human-computer interaction. Thus these approaches are too ``expensive'' to apply to even a moderate number of diverse items such as the PSTRE items in PIAAC 2012. \citet{he2016analyzing} adopted the concept of n-grams from natural language processing to explore the association between action sequence patterns and traditional item responses. The sequence patterns extracted from their procedure depend on the coding of log files and are often of limited capacity since it only considers consecutive actions.



In this paper, we propose a generic method to extract features from process data. The extracted features play a similar role as the latent variables in item response theory \citep{lord1980applications,lord1968statistical}. The proposed method does not rely on prior knowledge of the items and  coding of the log files. Therefore, it is applicable to a wide range of process data with little item-specific processing effort. In the case study,  we applied the proposed method to 14 PSTRE items in PIAAC 2012. These items vary widely in many aspects including the content of the problem-solving task and their overall difficulty levels.

The main component of the proposed feature extraction method is an autoencoder \citep[Chapter 14]{Goodfellow-et-al-2016}. It is a class of artificial neural networks that tries to reproduce the input in its output. Autoencoders are often used for dimension reduction \citep{hinton2006reducing} and data denoising \citep{vincent2008extracting} in pattern recognition, computer vision, and many other machine learning applications \citep{deng2010binary,lu2013speech,li2015hierarchical,yousefi2017autoencoder}. 
They first map the input to a low-dimensional vector, from which they then try to reconstruct the input. Once a good autoencoder is found for a dataset, the low-dimensional vector contains comprehensive information of original data and thus can be used as features summarizing the response processes.  

With the proposed method, we extract features from each of the PSTRE items in PIAAC 2012 and explore the extracted feature space of process data. We show that the extracted features from response processes contain more information than the traditional item responses. We find that the prediction of many variables, including literacy and numeracy scores and a variety of background variables, can be considerably improved once the process features are incorporated. 


Neural networks have been used for analyzing educational data recently. \citet{NIPS2015_5654} and \citet{wang2017deep} applied recurrent neural networks to knowledge tracing and showed that their deep knowledge tracing models can predict students' performance on the next exercise from their exercise trajectories more accurately than other traditional methods.  
\citet{bosch2017unsupervised} discussed several neural network architectures that can be used for analyzing interaction log data. They extracted features for detecting student boredom through modeling the relations of student behaviors in two time intervals. The log file data used there were aggregated into a more regular form. \citet{ding2019effective} also studied the problem of extracting features from student's learning process using autoencoders. The learning processes considered there have a fixed number of steps and the data in each step were preprocessed into fixed dimension raw features.

The rest of the paper is organized as follows. In Section \ref{sec:autoencoder}, we introduce the action sequence autoencoder and the feature extraction procedure for process data. The proposed procedure is applied to simulated processes in Section \ref{sec:simulation} to demonstrate how extracted features reveal the latent structure in response processes. Section \ref{sec:example} presents a case study of process data from PSTRE items of PIAAC to show that response processes contain more information than traditional responses. Some concluding remarks are made in Section \ref{sec:discussion}.

\section{Feature Extraction by Action Sequence Autoencoder}\label{sec:autoencoder}

We adopt the following setting throughout this paper. Let $\mathcal{A}=\{a_1, \ldots, a_N\}$ denote the set of possible actions for an item, where $N$ is the total number of distinct actions and each element in $\mathcal A$ is a unique action. A response process can be represented as a sequence of actions, $\bm s = (s_1, \ldots, s_{T})$, where $s_t \in \mathcal{A}$ for $t = 1, \ldots, T$ and $T$ denotes the length of the process, i.e., the total number of actions that a respondent took to solve the problem.
An action sequence $\bm s$ can be equivalently represented as a $T \times N$ binary matrix $\mathbf{S} = (S_{tj})$ whose $t$-th row gives the dummy variable representation of the action at time step $t$. More specifically, $S_{tj}$ being one indicates the $t$-th action of the sequence is action $a_j$. There is one and only one element being one in each row. All other elements are zeros. In the rest of this article, $\mathbf{S}$ is used interchangeably with $\bm{s}$ for referring to an action sequence.


The length of a response process is likely to vary widely across respondents. As a result, the matrix representation of response processes from different respondents will have different number of rows. For a set of $n$ processes, $\bm s_1, \ldots, \bm s_n$ (equivalently, $\mathbf{S}_1, \ldots, \mathbf{S}_n$), the length of $\bm s_i$ (the number of rows in $\mathbf{S}_i$) is denoted by $T_i$, for $i = 1, \ldots, n$. The main motivation of developing a feature extraction method for process data is to compress the nonstandard data with varying dimension into homogeneous dimension vectors to facilitate subsequent standard statistical analysis.

\subsection{Autoencoder}
\begin{figure}[htb]
\centering
\includegraphics[width=10cm]{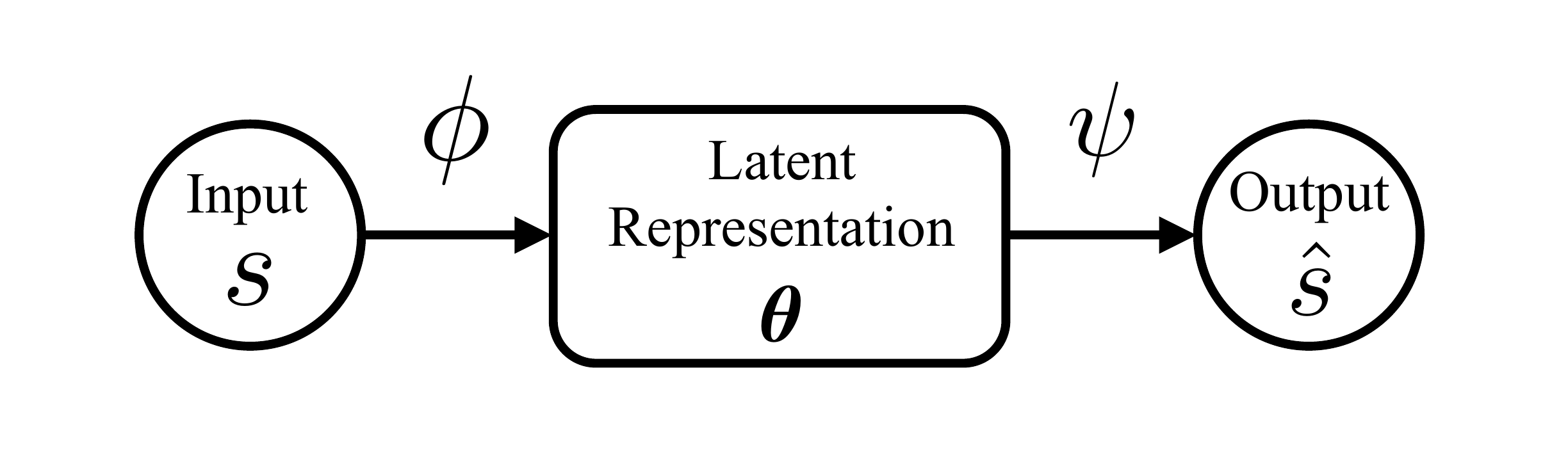}
\caption{Structure of an autoencoder.}\label{fig:autoencoder}
\end{figure}

The main component of our feature extraction method is an autoencoder \citep[Chapter 14]{Goodfellow-et-al-2016}. It is a type of artificial neural networks whose output tries to reproduce the input. A trivial solution to this task is to link the input and the output through an identity function, but it provides little insight about the data. Autoencoders employ special structures in the mapping from the input to the output so that nontrivial reconstructions are formed to unveil the underlying low-dimensional structure.
As illustrated in Figure \ref{fig:autoencoder},
an autoencoder consists of two components, an encoder $\phi$ and a decoder $\psi$. The encoder $\phi$ transforms a complex and high-dimensional input $\bm s$ into a low-dimensional vector $\bm \theta$. Then the decoder $\psi$ reconstructs the input from $\bm \theta$. Since the low-dimensional vector is in a standard and simpler format and contains adequate information to restore the original data, autoencoders are often used for dimension reduction and feature extraction.

The encoder and the decoder are often specified as a family of functions, $\phi_{\bm \eta}$ and $\psi_{\bm \xi}$, respectively, where $\bm \eta$ and $\bm \xi$ are parameters to be estimated by minimizing the discrepancy between the inputs and the outputs of the autoencoder. To be more specific, letting $\hat{\bm s}_i = \psi_{\bm \xi}(\phi_{\bm \eta}(\bm s_i))$ denote the output for input $\bm s_i$, $i = 1, \ldots, n$, the parameters $\bm \eta$ and $\bm \xi$ are estimated by minimizing 
\begin{equation}\label{eq:obj}
F(\bm \eta, \bm \xi) = \sum_{i=1}^n L(\bm s_i, \hat{\bm s}_i),
\end{equation} where $L$ is a loss function measuring the difference between the reconstructed data $\hat{\bm s}_i$ and the original data $\bm s_i$. 
Once estimates $\hat{\bm \eta}$ and $\hat{\bm \xi}$ are obtained, the latent representation or the features of an action sequence $\bm s$ can be computed by $\bm \theta = \phi_{\hat {\bm \eta}}(\bm s)$.

To make an analogue to the IRT models or other latent variable models, one may consider $\bm \theta$, the output of the encoder $\phi$, to be an estimator of the latent variables based on the responses and the decoder $\psi$ to be the item response function that specifies the response distribution corresponding to a latent vector. For the IRT model, the estimator and the item response function are often coherent in the sense that the estimator is determined by the item response function. For autoencoder, both $\phi$ and $\psi$ are parameterized and estimated based on the data. There is no coherence guarantee between them. This is one of the theoretical drawbacks of autoencoder. Nonetheless, we hope that the parametric families for $\phi$ and $\psi$ are flexible enough such that they can be consistently estimated with large samples and thus approximate coherence is automatically achieved.

Based on the above discussion, a crucial step in the application of autoencoders is to specify an encoder and a decoder that are suitable for the data to be compressed. In the remainder of this section, we will describe an autoencoder that performs well for response processes.
 
\subsection{Recurrent Neural Network}\label{sec:rnn}

\begin{figure}[htb]
\centering
\includegraphics[width=10cm]{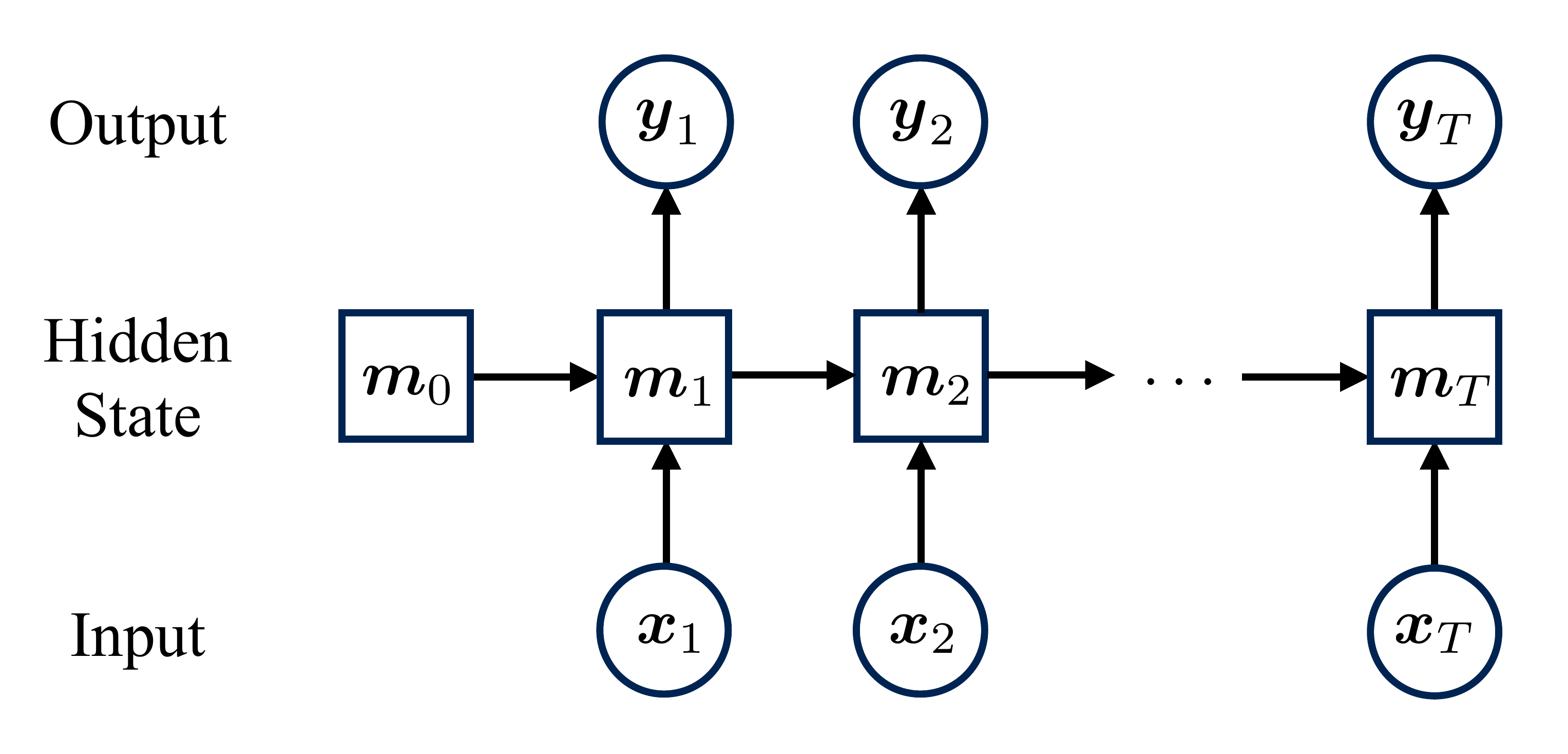}
\caption{Structure of RNNs.}\label{fig:rnn}
\end{figure}


To facilitate the presentation, we first provide a brief introduction to the recurrent neural networks (RNNs), a pivotal component of the encoder and the decoder of the action sequence autoencoder.

RNNs form a class of artificial neural networks that deal with sequences. Unlike traditional artificial neural networks such as multi-layer feed-forward networks \citep[Chapter 2]{patterson2017deep} that treat an input as a simple vector, RNNs have a special structure to utilize the sequential information in the data. 
As depicted in Figure \ref{fig:rnn}, the basic structure of RNNs has three components: inputs, hidden states, and outputs, each of which is a multivariate time series.
	The inputs $\bm x_1, \ldots, \bm x_T$ are $K$-dimensional vectors.
	The hidden states $\bm m_1, \ldots, \bm m_T$ are also $K$-dimensional and can be viewed as the memory that helps process the input information sequentially. The hidden state evolves as the input evolves.
	Each $\bm m_t$ summarizes what has happened up to time $t$ by integrating the current information $\bm x_t$ with the previous memory $\bm m_{t-1}$, that is, $\bm m_t$ is a function of $\bm x_t$ and $\bm m_{t-1}$
\begin{equation}\label{eq:rnn_hidden}
\bm m_t = f(\bm x_t, \bm m_{t-1}),
\end{equation}
for $t= 1, \ldots, T$. 
The initial hidden state $\bm m_0$ is often set to be the zero-vector. To extract from memory the information that is useful for subsequent tasks, a $K$-dimensional output vector $\bm y_t$ is produced as a function of the hidden state $\bm m_t$ at each time step $t$,
\begin{equation}\label{eq:rnn_output}
\bm y_t = g(\bm m_t).
\end{equation}
Both $f$ and $g$ are often specified as a parametric family of functions with parameters to be estimated from data. 

To summarize, an RNN makes use of the current input $\bm x_t$ and a summary of previous information $\bm m_{t-1}$ to produce an updated summary information $\bm m_t$, which in turn produces an output $\bm y_t$ at each time step $t$. 
An RNN is not a probabilistic model. It does not specify the probability distribution of the input $\bm x_t$ or the output $\bm y_t$ given the hidden state $\bm m_t$. It is essentially a deterministic nonlinear function that takes a sequence of vectors and outputs another sequence of vectors. Each output vector summarizes the useful information in the input vectors up to the current time step. We will write the function induced by an RNN as $\mathcal{R}(\cdot ; \bm \gamma)$ where $\bm \gamma$ collects the parameters in $f$ and $g$. Letting $\mathbf{X} = (\bm x_1, \ldots, \bm x_T)^\top$ and $\mathbf{Y} = (\bm y_1, \ldots, \bm y_T)^\top$ respectively denote the inputs and the outputs of the RNN, we have $\mathbf{Y} = \mathcal{R}(\mathbf{X}; \bm \gamma)$. We use a subscript $t$ of $\mathcal{R}$ to denote the output vector at time step $t$, that is, $\bm y_t = \mathcal{R}_t(\mathbf{X}; \bm \gamma)$.


RNNs can process sequences of different lengths. Note that the functions $f$ and $g$ in \eqref{eq:rnn_hidden} and \eqref{eq:rnn_output} are the same across time steps. Therefore, the total number of parameters for an RNN does not depend on the number of the time steps.

Various choices of $f$ and $g$ have been proposed to compute the hidden states and the outputs. Two most widely used ones are the long-short-term-memory (LSTM) unit \citep{hochreiter1997long} and the gate recurrent unit (GRU) \citep{cho2014learning}. They are designed to mitigate the vanishing or exploding gradient problem of a basic RNN \citep{bengio1994learning}. We will also use the two designs in the RNN component of our action sequence autoencoder. The detailed expressions of the LSTM unit and GRU are given in the appendix.


\subsection{Action Sequence Autoencoder}

\begin{figure}[htb]
\centering
\includegraphics[width=14cm]{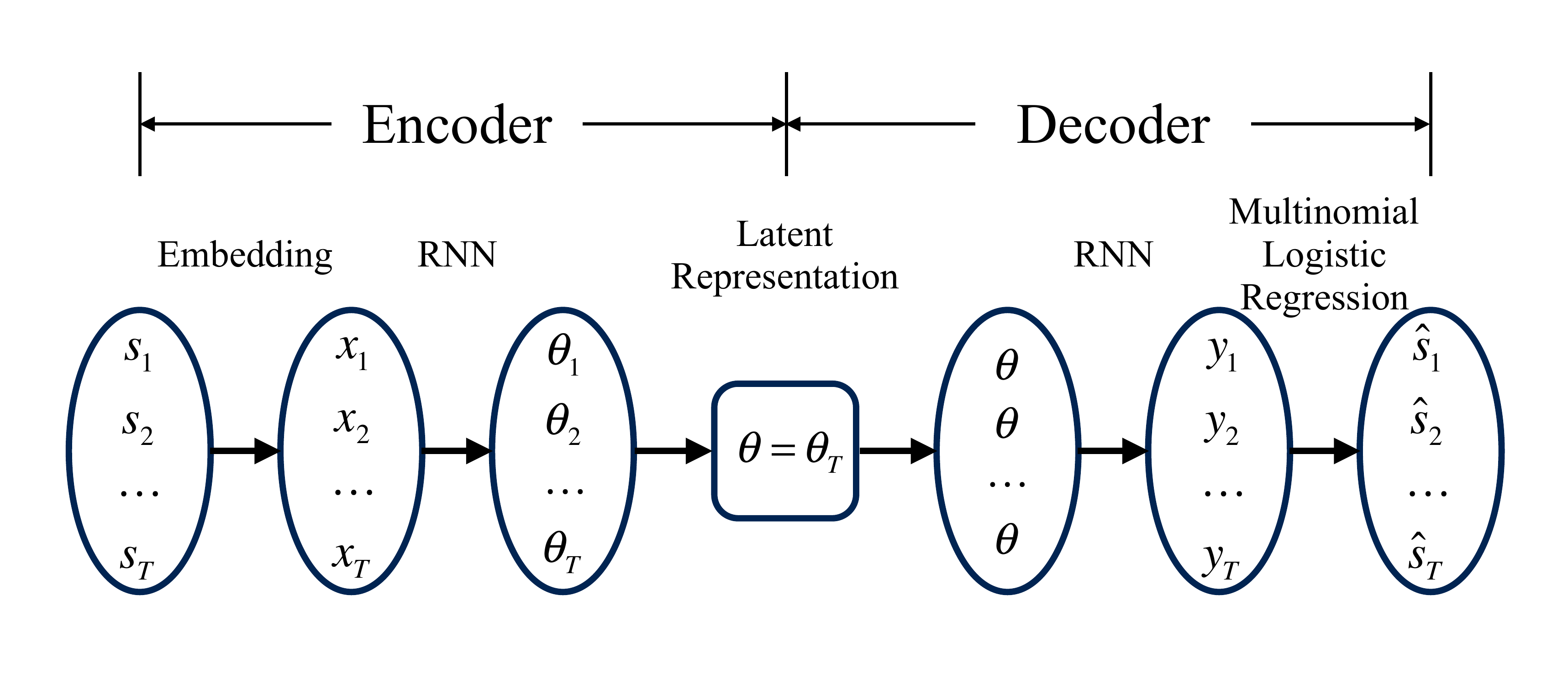}
\caption{Structure of action sequence autoencoders.}\label{fig:act_seq_autoencoder}
\end{figure}

The action sequence autoencoder used for extracting features from process data takes a sequence of actions as the input process and outputs a reconstructed sequence.
The diagram in Figure \ref{fig:act_seq_autoencoder} illustrates the structure of the action sequence autoencoder. In what follows, we elaborate the encoding and the decoding mechanism.

\paragraph{Encoder.} The encoder of the action sequence autoencoder takes a sequence of actions and outputs a $K$-dimensional vector as a compressed summary of the input action sequence. Working with action sequences directly is often challenging because of the categorical nature of the actions.
To overcome the obstacle, we associate each action $a_i$ in the action pool $\mathcal{A}$ with a $K$-dimensional latent vector $\bm e_i$ that will be estimated based on the data. These latent vectors describe the attributes of actions and will be used to summarize the information contained in the sequence. The method of mapping categorical variables to continuous latent attributes is often called the embedding method. It is widely used in machine learning applications such as neural machine translation and knowledge graph completion \citep{bengio2003neural,NIPS2013_5021,kraft2016embedding}.

The first operation of our encoder is to transform the input sequence $\bm s=(s_1, \ldots, s_T)$ into a corresponding sequence of latent vectors $(\bm e_{i_1}, \ldots, \bm e_{i_T})$ where $i_t$ is the index of the action in $\mathcal{A}$ at time step $t$, that is, $s_t = a_{i_t}$ for $t = 1, \ldots, T$. 
With the binary matrix representation $\mathbf{S}$ of action sequence $\bm s$, the embedding step of the encoder is simply a matrix multiplication $\mathbf{X} = \mathbf{SE}$ where $\mathbf{E} = (\bm e_1, \ldots, \bm e_N)^\top$ is an $N \times K$ matrix whose $i$-th row is the latent vector for action $a_i$ and the rows of $\mathbf{X} = (\bm e_{i_1}, \ldots, \bm e_{i_T})^\top$ form the latent vector sequence corresponding to the original action sequence $\bm s$.

Given the latent vector sequence, the encoder uses an RNN to summarize the information. 
Since our goal is to compress the entire response process into a single $K$-dimensional vector, only the last output vector of the RNN is kept to serve as a summary of information. Therefore, the output of the encoder, i.e., the latent representation of the input sequence, is $\bm \theta = \mathcal{R}_T(\mathbf{X}; \bm \gamma_{\text{E}})$. 

To summarize, the encoder of our action sequence autoencoder is
\begin{equation}\label{eq:encoder}
\phi_{\bm \eta}(\mathbf{S}) = \mathcal{R}_T(\mathbf{SE}; \bm \gamma_{\text{E}}), 
\end{equation}
where $\bm \eta$ represents all the parameters including the embedding matrix $\mathbf{E}$ and the parameter vector $\bm \gamma_{\text{E}}$ of the encoder RNN. The encoding procedure consists of the following three steps. 
\begin{enumerate}
\item An observed action sequence is transformed into a sequence of latent vectors by the embedding method: $\mathbf{X} = \mathbf{S}\mathbf{E}$.
\item The latent vector sequence is processed by the encoder RNN to obtain another sequence of vectors $\mathcal{R}(\mathbf{X}; \bm \gamma_{\text{E}}) = (\bm \theta_1, \ldots, \bm \theta_T)^\top$ where $\bm \theta_t = \mathcal{R}_t(\mathbf{X}; \bm \gamma_{\text{E}})$ for $t=1,\ldots, T$.
\item The last output of the RNN is kept as the latent representation, namely, $\bm \theta = \bm \theta_T$.
\end{enumerate}
Each of the three steps corresponds to an arrow in the encoder part of Figure \ref{fig:act_seq_autoencoder}.

%
%

\paragraph{Decoder.} The decoder of the action sequence autoencoder reconstructs an action sequence $\bm s$, or equivalently, its binary matrix representation $\mathbf{S}$, from $\bm \theta$.
First, a different RNN is used to expand the latent representation $\bm \theta$ into a sequence of vectors, each of which contains the information of the action at the corresponding time step. As $\bm \theta$ is the only information available for the reconstruction, the input of the decoder RNN is the same $\bm \theta$ for each of the $T$ time steps. Writing it in a matrix form, the input of the decoder RNN is $\mathbf{1}_T \bm \theta^\top$ where $\bm{1}_T$ is the $T$-dimensional vector of ones. After the decoder RNN's processing, we obtain a sequence of $K$-dimensional vectors $\mathbf{Y} = (\bm y_1, \ldots, \bm y_T)^\top = \mathcal{R}(\bm{1}_T{\bm\theta}^\top; \bm \gamma_\text{D})$. Each $\bm y_t$ contains the information for the action taken at time step $t$.
	
Recall that each row of $\mathbf{S}$ is the dummy variable representation of the action taken at corresponding time step.
Each row essentially specifies a degenerate categorical distribution on $\mathcal{A}$, with the action that is actually taken having probability one and all the other actions having probability zero. 
With this observation, the task of restoring the action at step $t$ becomes constructing the probability distribution of the action taken at step $t$ from $\bm y_t$. 
The multinomial logit model (MLM) can be used in the decoder to achieve this. To be more specific, the probability of taking action $a_j$ at time $t$ is 
\begin{equation}\label{eq:mlm}
\hat{S}_{tj} = \left\{
\begin{array}{ll}
\frac{\exp(b_j + \bm y_t^\top \bm \beta_j)}{ 1 + \sum_{k=1}^{N-1} \exp(b_k + \bm y_t^\top \bm \beta_k)} & \text{if} ~j= 1, \ldots, N-1;\\
\frac{1}{ 1 + \sum_{k=1}^{N-1} \exp(b_k + \bm y_t^\top \bm \beta_k)} & \text{if}~j=N,
\end{array}\right.
\end{equation}
where $b_j$ and $\beta_j$ are parameters to be estimated from the data. Note that the parameters in \eqref{eq:mlm} do not depend on $t$. That is, the encoder uses the same MLM to compute the probability distribution of $s_t$ from $\bm y_t$ for $t=1, \ldots, T$.
As a result, the reconstructed sequence is $\hat{\mathbf{S}} = (\hat{S}_{tj})$ and the decoder can be written as
\begin{equation}\label{eq:decoder}
\psi_{\bm \xi}(\bm \theta) = \text{MLM}(\mathcal{R}(\bm{1}_T{\bm\theta}^\top; \bm \gamma_\text{D})),
\end{equation}
where the parameter vector $\bm \xi$ consists of the parameter vector $\bm \gamma_{\text{D}}$ in the decoder RNN and $b_j, \bm \beta_j$, $j = 1, \ldots, N-1$.

If we have an ideal autoencoder that reconstructs the input perfectly, the probability distribution specified by $(\hat{S}_{t1}, \ldots, \hat{S}_{tN})$ will concentrate all its probability mass on the action that is actually taken. In practice, it is very unlikely to construct such perfect autoencoders. Usually, every action in the action set $\mathcal{A}$ will be assigned a positive probability in the reconstructed probability distribution. For a given set of response processes, we want to manipulate the parameters in the encoder and the decoder so that the reconstructed probability distribution concentrates as much probability mass on the actual action as possible.

To summarize, as depicted in the decoder part of Figure \ref{fig:act_seq_autoencoder}, the decoding procedure of the action sequence autoencoder consists of the following three steps.
\begin{enumerate}
\item The latent representation $\bm \theta$ is replicated $T$ times to form the $T \times K$ matrix $\mathbf{1}_T \bm \theta^\top$.
\item The decoder RNN takes $\mathbf{1}_T \bm \theta^\top$ and outputs a sequence of vectors $(\bm y_1, \ldots, \bm y_T)$, each element of which containing the information of the action at the corresponding step.
\item The probability distribution of $s_t$ is computed according to the MLM from $\bm y_t$ at each time step $t$.
\end{enumerate}

\paragraph{Loss function.}
In order to extract good features for a given set of response processes, we need to construct an action sequence autoencoder that reconstructs the response processes as well as possible. The discrepancy between an action sequence $\mathbf{S}$ and its reconstructed version $\hat{\mathbf{S}}$ can be measured by the following loss function
\begin{equation} \label{eq:cross_entropy}
L(\mathbf{S}, \hat{\mathbf{S}}) = -\frac{1}{T}\sum_{t=1}^T \sum_{j=1}^N S_{tj} \log(\hat{S}_{tj}).
\end{equation}
Note that, for a given $t$, only one of $S_{t1}, \ldots, S_{tN}$ is non-zero. The loss function is smaller if the distribution specified by $(\hat{S}_{t1}, \ldots, \hat{S}_{tN})$ is more concentrated on the action that is actually taken at step $t$. The best action sequence autoencoder for describing a given set of response processes is the one that minimizes the total reconstruction loss defined in \eqref{eq:obj}.

Notice that \eqref{eq:cross_entropy} is in the same form as the log-likelihood function of categorical distributions. By using this loss function, we implicitly define a probabilistic model for the response processes. That is, given the latent representation $\bm \theta$, $s_t$ follows a categorical distribution on $\mathcal{A}$ with probability vector $(\hat{S}_{t1}, \ldots, \hat{S}_{tN})$. The decoder of the action sequence autoencoder specifies the functional form of the probability vector in terms of $\bm \theta$ and $\bm \xi$.


\subsection{Procedure}
Based on the above discussion, we extract $K$ features from $n$ response processes $\mathbf{S}_1, \ldots, \mathbf{S}_n$ through the following procedure.
\begin{procedure}[Feature extraction for process data]\label{proc:feature}~
\begin{enumerate}
\item Find a minimizer, $(\hat{\bm \eta}, \hat{\bm \xi})$, of the objective function $F(\bm \eta, \bm \xi) = \sum_{i=1}^n L(\mathbf{S}_i, \hat{\mathbf{S}}_i)$ by stochastic gradient descent through the following steps.
\begin{enumerate}

\item Initialize the parameters $\bm \eta$ and $\bm \xi$.
\item Randomly generate $i$ from $\{1, \ldots, n\}$ and update $\bm \eta$ and $\bm \xi$ with $\bm \eta - \alpha \frac{\partial{L(\mathbf{S}_i, \hat{\mathbf{S}}_i})}{\partial \bm\eta}$ and $\bm \xi - \alpha \frac{\partial{L(\mathbf{S}_i, \hat{\mathbf{S}}_i})}{\partial \bm\xi}$, respectively, where $\alpha$ is a predetermined small positive number.\label{step2}
\item Repeat step (b) until convergence.
\end{enumerate}



\item Calculate $\tilde{\bm\theta}_i = \phi_{\hat{\bm\eta}}(\mathbf{S}_i)$, for $i=1, \ldots, n$. Each column of $\tilde{\bm\Theta} = (\tilde{\bm \theta}_1, \ldots, \tilde{\bm\theta}_n)^\top$ is a raw feature of the response processes.
\item Perform principal component analysis (PCA) on $\tilde{\bm\Theta}$. The principal components are the $K$ principal features of the response processes.
\end{enumerate}
\end{procedure}

In Step 1, the optimization problem is solved by stochastic gradient descent (SGD) \citep{robbins1951stochastic}. In Step 1b, a fixed step size $\alpha$ is used for updating the parameters. Data-dependent step sizes such as those proposed in \citet{duchi2011adaptive}, \citet{zeiler2012adadelta}, and \citet{kingma2014adam} can be easily adapted for the optimization problem.

Neural networks are often over-parametrized. To prevent overfitting, validation based early stopping \citep{Prechelt2012} is often used when estimating parameters of complicated neural networks such as our action sequence autoencoder. 
With this technique, the optimization algorithm, in our case, SGD, is not run until convergence. A parameter value that are obtained before convergence with good performance on the validation set is used as an estimate of the minimizer.
To perform early stopping, a dataset is split into a training set and a validation set. A chosen optimization algorithm is performed only on the training set for a number of epochs. An epoch consists of $n_\text{T}$ iterations, where $n_\text{T}$ is the size of the training set. At the end of each epoch, the objective function is evaluated on the validation set. The value of the parameters produces the lowest validation loss is used as an estimate of the minimizer. 
We adopt this technique when constructing the action sequence autoencoder. The feature extraction procedure with validation-based early stopping is summarized in  Procedure \ref{proc:feature_valid}.

\begin{procedure}[Feature extraction with validation-based early stopping]\label{proc:feature_valid}~
\begin{enumerate}
\item Find a minimizer, $(\hat{\bm \eta}, \hat{\bm \xi})$, of the objective function $F(\bm \eta, \bm \xi) = \sum_{i=1}^n L(\mathbf{S}_i, \hat{\mathbf{S}}_i)$ by stochastic gradient descent with validation-based early stopping through the following steps.
\begin{enumerate}
	\item Randomly split $\{1, \ldots, n\}$ into a training index set $\Omega_\text{T}$ of size $n_\text T$ and a validation index set $\Omega_\text{V}$ of size $n_\text V$.
	\item Initialize the parameters $\bm \eta$ and $\bm \xi$ and calculate $F_{\text{V}1} =  \sum_{i \in \Omega_{\text V}} L(\mathbf{S}_i, \hat{\mathbf{S}}_i)$.
	\item Randomly permute the indices in $\Omega_T$ and denote the result as $(i_1, \ldots, i_{n_{\text T}})$.
	\item For $k = 1, \ldots, n_{\text T}$, update $\bm \eta$ and $\bm \xi$ with $\bm \eta - \alpha \frac{\partial{L(\mathbf{S}_{i_k}, \hat{\mathbf{S}}_{i_k})}}{\partial \bm\eta}$ and $\bm \xi - \alpha \frac{\partial{L(\mathbf{S}_{i_k}, \hat{\mathbf{S}}_{i_k})}}{\partial \bm\xi}$, respectively.
	\item Calculate $F_{\text{V}2} =  \sum_{i \in \Omega_{\text V}} L(\mathbf{S}_i, \hat{\mathbf{S}}_i)$. If $F_{\text{V}2}$ is smaller than $F_{\text{V}1}$, let $\hat{\bm \eta} = \bm \eta$ and $\hat{\bm \xi} = \bm \xi$ and update $F_{\text{V}1}$ with $F_{\text{V}2}$.
\item Repeat steps (c), (d), and (e) for sufficiently many times.
\end{enumerate}

\item Calculate $\tilde{\bm\theta}_i = \phi_{\hat{\bm\eta}}(\mathbf{S}_i)$, for $i=1, \ldots, n$. Each column of $\tilde{\bm\Theta} = (\tilde{\bm \theta}_1, \ldots, \tilde{\bm\theta}_n)^\top$ is a raw feature of the response processes.

\item Perform principal component analysis (PCA) on $\tilde{\bm\Theta}$. The principal components are the $K$ principal features of the response processes.
\end{enumerate}
\end{procedure}

The proposed feature extraction procedure requires the number of features to be extracted, $K$, as an input. In general, if $K$ is too small, the action sequence autoencoder does not have enough flexibility to capture the structure of the response processes. On the other hand, if $K$ is too big, the extracted features contain too much redundant information, causing overfitting and instability in downstream analyses. We adopt the $k$-fold cross-validation procedure \citep{stone1974cross} to choose a suitable $K$ in the analyses presented in Sections \ref{sec:simulation} and \ref{sec:example}.

We perform principal component analysis on the raw features in the last step of the proposed feature extraction procedure for seeking for feature interpretations. As we will show in the case study, the first several principal features usually have clear interpretations even if the meaning of the actions is not taken into account in the feature extraction procedure.

Since the extracted features have a standard format, they can be easily incorporated in (generalized) linear models and many other well-developed statistical procedures.
As we will show in the sequel, the extracted features contain a substantial amount of information about the action sequences. They can be used as surrogates of the action sequences to study how response processes are related to the respondents' latent traits and other quantities of interest.



\section{Simulations}\label{sec:simulation}
\subsection{Experiment Settings}
In this section, we apply the proposed feature extraction method to simulated response processes of an item with 26 possible actions. Each action in the item is denoted by an upper-case English letter. In other words, we define $\mathcal{A} = \{\text{A}, \text{B}, \ldots, \text{Z}\}$. All the sequences used in the study start with A and end with Z, meaning that A and Z represent the start and the end of an item, respectively. 

In our simulation study, action sequences are generated from Markov chains. That is, given the first $t$ actions in a response process, $s_1, \ldots, s_t$, the distribution from which $s_{t+1}$ is generated depends only on $s_t$. 
A Markov chain is determined by its probability transition matrix $\mathbf{P} = (p_{ij})_{1\leq i, j \leq N}$, where $p_{ij} = P(s_{t+1} = a_j \, | \, s_{t} = a_i)$. Because of the special meaning of actions A and Z, there should not be transitions from other actions to A and from Z to other actions. As a result, the probability transition matrices used in our simulation study have the constraints that $p_{i1} = 0$ for $i = 1, \ldots, N$, and $p_{NN} = 1$. To construct a probability transition matrix, we only need to specify its elements in its upper right $(N-1)\times (N-1)$ submatrix. Given a transition matrix $\mathbf{P}$, we start a sequence with A and generate all subsequent actions according to $\mathbf{P}$ until Z appears.

Two simulation scenarios are devised in our experiments to impose latent class structures in generated response processes. In Scenario I, two latent groups are formed by generating action sequences from two different Markov chains. Let $\mathbf{P}_1$ and $\mathbf{P}_2$ denote the probability transition matrices of the two chains. A set of $n$ sequences are obtained by generating $n/2$ sequences according to $\mathbf{P}_1$ and the remaining $n/2$ sequences according to $\mathbf{P}_2$. Both $\mathbf{P}_1$ and $\mathbf{P}_2$ are randomly generated and then fixed to generate all sets of response processes. To generate $\mathbf{P}_1 = (p_{ij}^{(1)})_{1\leq i,j \leq N}$, we first construct an $(N-1) \times (N-1)$ matrix $\mathbf{U}$ whose elements are independent samples from a uniform distribution on interval $[-10, 10]$. Then the upper right $(N-1)\times (N-1)$ submatrix of $\mathbf{P}_1$ is computed from $\mathbf{U}$ by
\begin{equation}
{p}_{i+1,j}^{(1)} = \frac{\exp(u_{ij})} { \sum_{l=1}^{N-1}\exp(u_{il})},~ i,j = 1, \ldots, N-1.
\end{equation}
The transition matrix $\mathbf{P}_2$ is obtained similarly.

In Scenario II, half of the $n$ action sequences in a set are generated from $\mathbf{P}_1$ as in Scenario I. The other half is obtained by reversing the actions between A and Z in each of the generated sequences. For example, if (A, B, C, Z) is a generated sequence, then the corresponding reversed sequence is (A, C, B, Z). The two latent groups formed in this scenario is more subtle than those in Scenario I as a sequence and its reversed version cannot be distinguished by marginal counts of actions in $\mathcal{A}$.


We consider three choices of $n$, 500, 1000, and 2000. One hundred sets of action sequences are generated for each simulation scenario and each choice of $n$. Procedure \ref{proc:feature_valid} is applied to each datasets. Both LSTM and GRU are considered for the recurrent unit in the autoencoder. For each choice of the recurrent unit, the number of features to be extracted are chosen from $\{10, 20, 30, 40, 50\}$ by five-fold cross-validation.

We investigate the ability of the extracted features in preserving the information in action sequences by examining their performance in reconstructing variables derived from action sequences. The variables to be reconstructed are indicators of the appearance of an action or an action pair in a sequence. Rare actions and action pairs that appears fewer than $0.05n$ times in a dataset are not taken into consideration. We model the relationship between the indicators and the extracted features through logistic regression. For each dataset, $n$ sequences are split into training and test sets in the ratio of 4:1. A logistic regression model is estimated for each indicator on the training set and its prediction performance is evaluated on the test set by the proportion of correct prediction, i.e., prediction accuracy. The average prediction accuracy over all the considered indicators are recorded for each dataset and each choice of the recurrent unit.

To study how well the extracted features unveil the latent group structures in response processes, we build a logistic regression model to classify the action sequences according to the extracted features. The training and test sets are split similarly as before and the prediction accuracy on the test set is recorded for evaluation.

\subsection{Results}
\begin{table}[htb]
\centering
\caption{Mean (standard deviation) of prediction accuracy in the simulation study.}\label{table:sim_ex1}
\begin{tabular}{ccccccc}
\hline
\multirow{2}{*}{Scenario} & \multirow{2}{*}{$n$} & \multicolumn{2}{c}{Reconstruction Accuracy} && \multicolumn{2}{c}{Group Accuracy} \\
\cline{3-4}
\cline{6-7}
 & & LSTM & GRU && LSTM & GRU \\
\hline
\multirow{3}{*}{I} & 500 & 0.88 (0.005) & 0.87 (0.006) && 0.99 (0.010) & 1.00 (0.007) \\
& 1000 & 0.90 (0.003) & 0.90 (0.004) && 0.99 (0.005) & 0.99 (0.006)\\
& 2000 & 0.91 (0.002) & 0.91 (0.003)&& 0.99 (0.005) & 0.99 (0.005)\\
\hline
\multirow{3}{*}{II} & 500 & 0.88 (0.006) & 0.88 (0.006) && 0.86 (0.033) & 0.87 (0.031) \\
& 1000 & 0.90 (0.004) & 0.91 (0.005) && 0.86 (0.021) & 0.86 (0.021)\\
& 2000 & 0.91 (0.002) & 0.92 (0.003) && 0.87 (0.027) & 0.87 (0.016)\\
\hline
\end{tabular}
\end{table}

Table \ref{table:sim_ex1} reports the results of our simulation study. A few observations can be made from Table \ref{table:sim_ex1}.
First, the accuracy for reconstructing the appearance of actions and action pairs is high in both simulation scenarios, indicating that the extracted features preserve a significant amount of information in the original action sequences. The reconstruction accuracy is slightly improved as $n$ increases. Including more action sequences can provide more information for estimating the autoencoder in Step 1 of Procedure \ref{proc:feature_valid} thus producing better features. A larger sample size can also lead to a better fit of the logistic models that relate features to derived variables. Both effects contribute to the improvement of action and action pair reconstruction.

\begin{figure}[htb]
\centering
\includegraphics[width=0.45\textwidth]{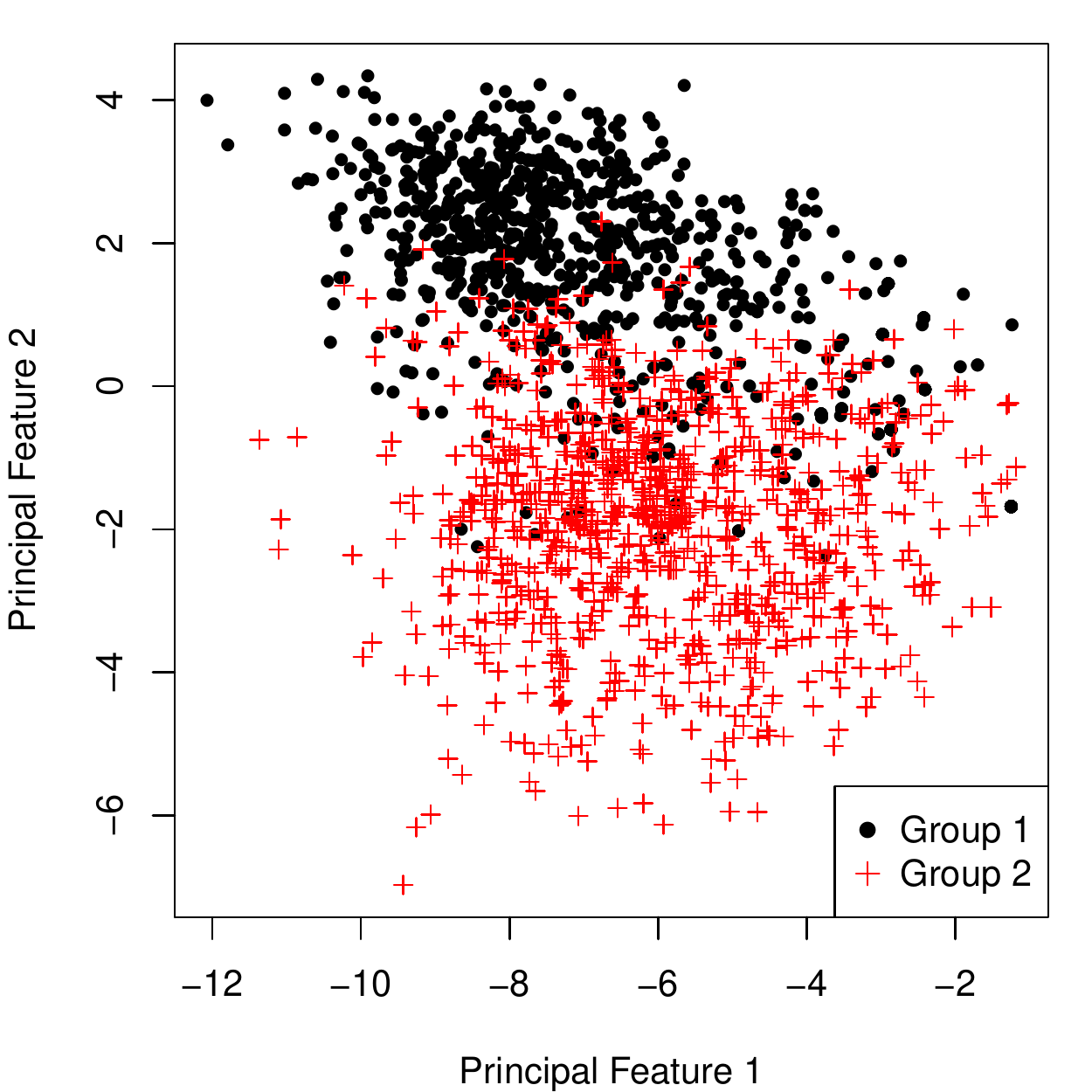}
~
\includegraphics[width=0.45\textwidth]{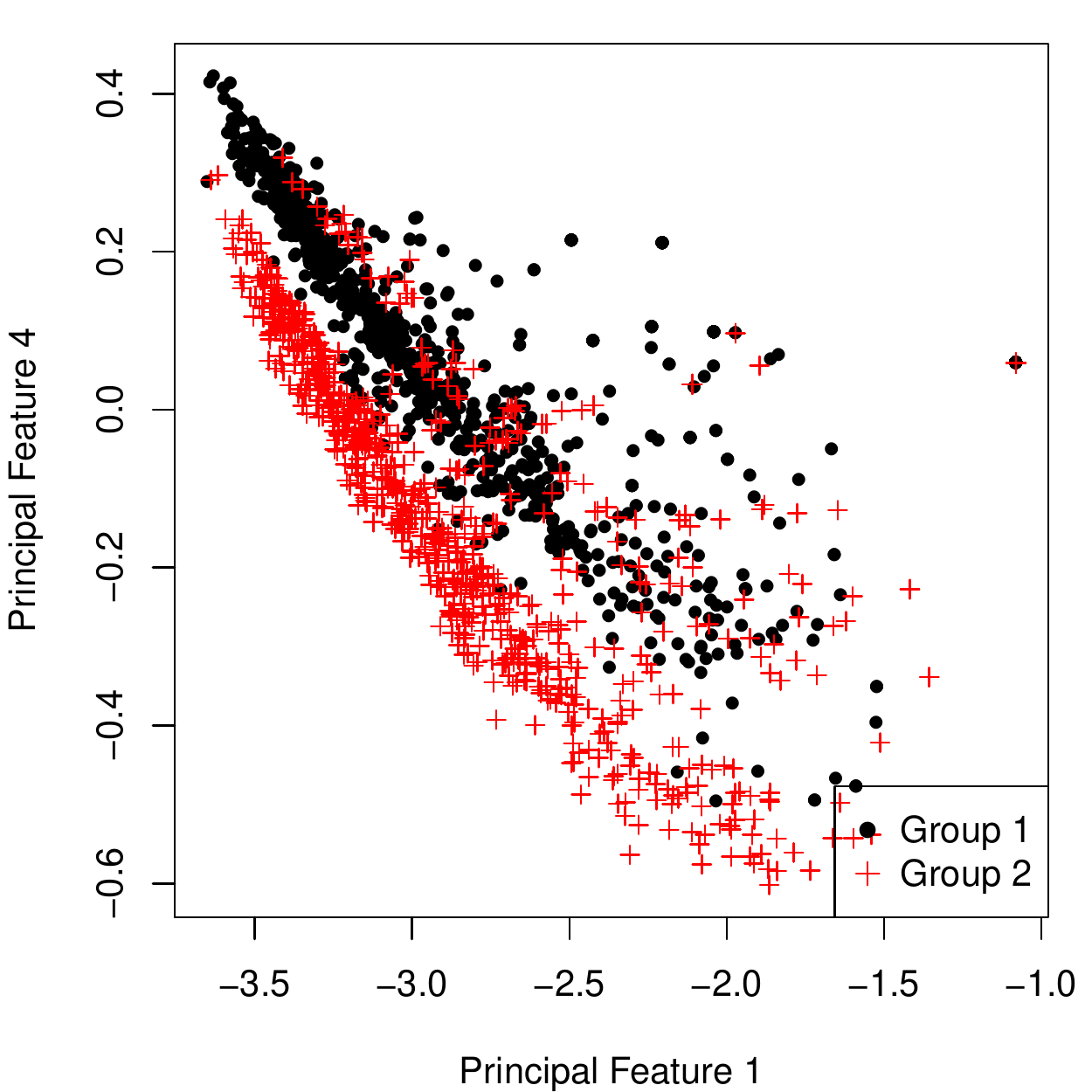}
\caption{Left: scatterplot of the first two principal features for one dataset of 2000 sequences generated under scenario I. Right: scatterplot of principal features 1 and 4 for one dataset of 2000 sequences generated under scenario II.}\label{fig:sim_latent}
\end{figure}

Second, in both simulation scenarios, the extracted features can distinguish the two latent groups well. In Scenario I, the two groups can be separated almost perfectly. Since the group difference in Scenario II is more subtle, the accuracy in classifying the two groups is lower than that in Scenario I, but still more than 85\% of the sequences can be classified correctly. To further look at how the extracted features unveil the latent structure of action sequences, we plot two principal features for one of the datasets with 2000 sequences for each scenario in Figure \ref{fig:sim_latent}. The left panel presents the first two principal features for Scenario I. The group structure is clearly shown and the two groups can be roughly separated by a horizontal line at 0. The right panel of Figure \ref{fig:sim_latent} displays the plot of the first and fourth principal features for Scenario II. Again the two groups can be clearly separated.

Last, the extracted features for the two choices of the recurrent unit in the action sequence autoencoder are comparable in terms of both reconstruction and group structure identification. A GRU has a simpler structure and fewer parameters than an LSTM unit with the same latent dimension. In this sense, GRU is more efficient for our action sequence modeling.

\section{Case Study}\label{sec:example}
\subsection{Data}
Process data used in this study contains 11,464 respondents' response processes of the PSTRE items in PIAAC 2012. There are 14 PSTRE items in total. In our data, 7,620 respondents answered 7 items and 3,645 respondents answered all 14 items. For each of the 14 items, there are around 7,500 respondents. For each respondent-item pair, both the response process (action sequence) and the final response outcome were recorded. The original final outcomes for some items are polytomous. We simplify them into binary outcomes with the fully corrected responses labelled as 1 and all others as 0.



The 14 PSTRE items in PIAAC 2012 vary in content, task complexity and difficulty. Some basic descriptive statistics of the items are summarized in Table \ref{table:items}, where $n$ denotes the number of respondents, $N$ is the number of possible actions, $\bar T$ stands for the average sequence length and Correct \% is the percentage of correct responses. There are three types of interaction environments, email client, spreadsheet, and, web browser. Some items such as U01a and U01b have a single environment while some items such as U02 and U23 involve multiple environments. U06a is the simplest item in terms of number of possible actions and average response length, but only about one fourth of the participants answered it correctly. Items U02 and U04a are the most difficult items---only around 10\% of the respondents correctly completed the given tasks. The tasks in these two items are relatively complicated---there are a few hundred of possible actions and more than 40 actions are needed to finish the task. With the wide item variety, manually extracting important features of process data based on experts' understanding of the items is time-consuming while the proposed automatic method can be easily applied to all these items.

\begin{table}[htb]
\caption{Descriptive statistics of PIAAC PSTRE items.}\label{table:items}
\centering
\begin{threeparttable}
\begin{tabular}{llccccc}
\hline
ID & Description & $n$ & $N$ & {\small $\bar T$} &Correct \%\\
\hline
U01a & Party Invitations - Can/Cannot Come & 7620 & 207 & 24.8 & 54.5\\
U01b & Party Invitations - Accommodations & 7670 & 249 & 52.9 & 49.3\\
U02 & Meeting Rooms & 7537 & 328 & 54.1 & 12.8\\
U03a & CD Tally & 7613 & 280 & 13.7 & 37.9\\
U04a & Class Attendance & 7617& 986 & 44.3 & 11.9\\
U06a & Sprained Ankle - Site Evaluation Table & 7622 & 47 & 10.8 & 26.4 \\
U06b & Sprained Ankle - Reliable/Trustworthy Site & 7612 & 98 & 16.0 & 52.3\\
U07 & Digital Photography Book Purchase & 7549 & 125 & 18.6 & 46.0\\
U11b & Locate E-mail - File 3 E-mails & 7528 & 236 & 30.9 & 20.1 \\
U16 & Reply All & 7531 & 257 & 96.9 & 57.0\\
U19a & Club Membership - Member ID & 7556 & 373 & 26.9 & 69.4\\
U19b & Club Membership - Eligibility for Club President & 7558 & 458 & 21.3 & 46.3\\
U21 & Tickets & 7606 & 252 & 23.4 & 38.2\\
U23 & Lamp Return & 7540 & 303 & 28.6 & 34.3\\
\hline
\end{tabular}
\begin{tablenotes}
\item Note: $n$ = number of respondents; $N$ = number of possible actions; $\bar T$ = average sequence length; Correct \% = percentage of correct responses.
\end{tablenotes}
\end{threeparttable}
\end{table} 


\subsection{Features}
We extract features from the response processes for each of the 14 items using the proposed procedure. The number of features is chosen from $\{10, 20, \ldots, 100\}$ by five-fold cross-validation. Adam \citep{kingma2014adam} step size is used for optimizing the object function in Step 1 of Procedure \ref{proc:feature_valid}. The algorithm is run for 100 epochs with validation based early stopping, where 10\% of the processes are randomly sampled to form the validation set for each item.

Although the proposed method does not utilize the meaning of the actions for feature extraction, many of the principal features, especially the first several, have clear interpretations. Table \ref{table:feature} gives a partial list of feature interpretations. 


\begin{table}[htb]
\centering
\caption{A partial list of feature interpretations by interface type.}\label{table:feature}
\begin{tabular}{cc}
\hline
Interface Type & Interpretation \\
\hline
\multirow{2}{*}{Email Client} & viewing emails and folders, moving emails,\\
 & creating new folders, typing emails\\
 \hline
\multirow{2}{*}{Spreadsheet} & using sort, using search, \\
 & clicking drop-down menu\\
 \hline
\multirow{2}{*}{Web Browser} & clicking relevant links,\\
 & clicking irrelevant links\\
\hline
\multirow{4}{*}{All Interfaces} & sequence length, \\
 & using actions related to the task,\\
 & switching working environments, \\
 & selecting answers, answer submission \\
\hline
\end{tabular}
\end{table}

The first or the second principal feature of each item is usually related to respondents' attentiveness. An inattentive respondent tends to move to the next item without meaningful interactions with the computer environment. In contrast, an attentive respondent typically tries to understand and to complete the task by exploring the environment. Thus attentiveness in response process can be reflected in the length of the process. We call the principal feature that has the largest absolute correlation with the logarithm of the process length the attentiveness feature. In our case, the attentiveness feature is the second principal feature for item U06a and the first for all other items. For all the items, the absolute correlation between the attentiveness feature and the logarithm of sequence length is higher than 0.85. To make a higher attentiveness feature correspond to a more attentive respondent, we modify the attentiveness features by multiplying each of them by the sign of their correlation with the logarithm of process length. 
For a given pair of items, we select respondents who responded to both items and calculate the correlation between the two modified attentiveness features. These correlations are all positive and range from 0.30 to 0.70, implying that the respondents who are inattentive in one item tend to be inattentive in another item.

\begin{figure}[htb]
\centering
\includegraphics[width=\textwidth]{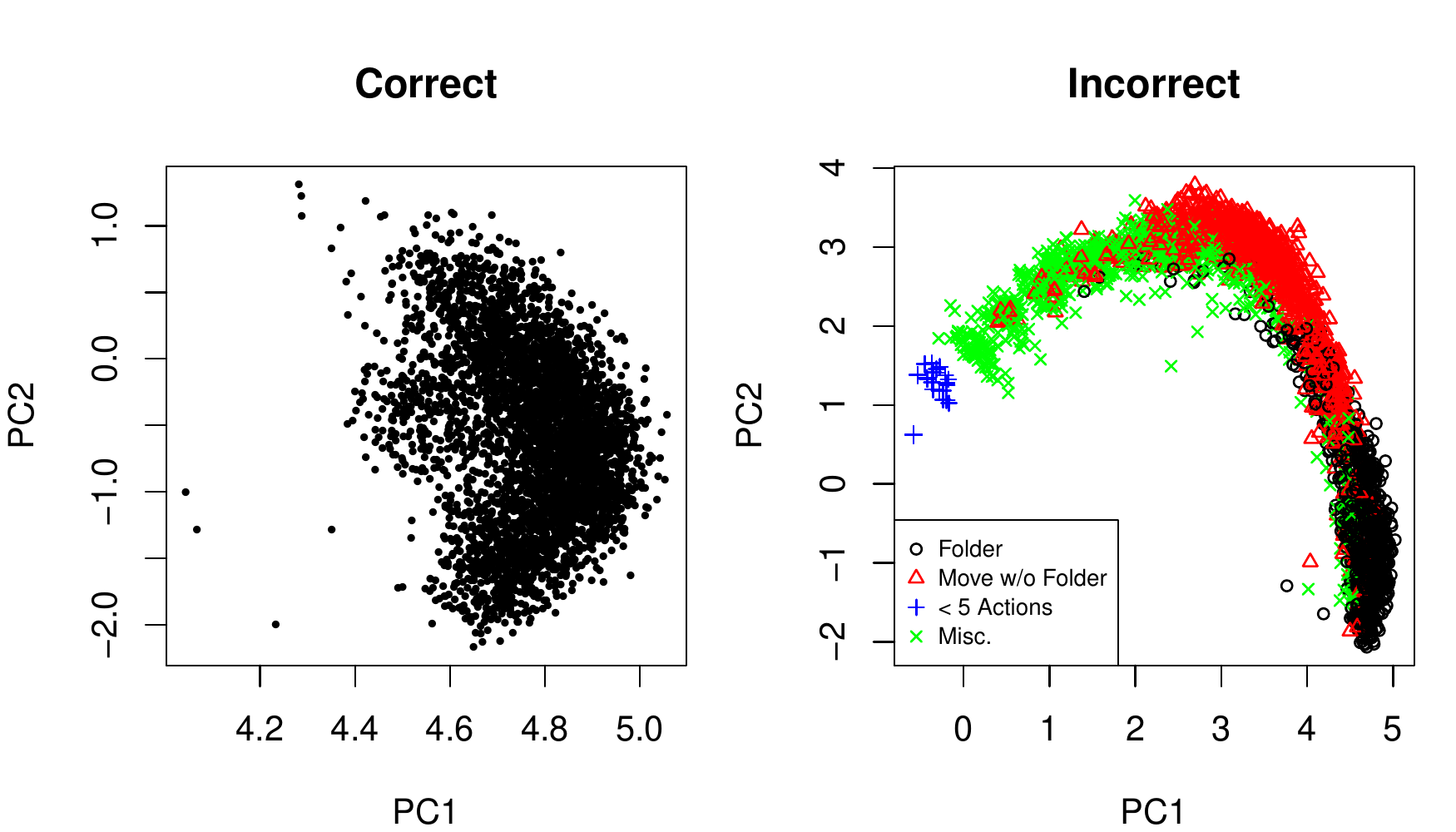}
\caption{Scatterplots of the first two principal features of U01b stratified by response outcome.}\label{fig:feature_U01b}
\end{figure}

The feature space of the respondents with correct responses is usually very different from that of the respondents with incorrect responses. As an illustration, in Figure \ref{fig:feature_U01b}, we plot the first two principal features of U01b for the two groups of respondents separately. It is obvious that the two clouds of points are of very distinct shapes. The non-oval shape of the clouds suggests that the feature space is highly non-linear. A multivariate normal distribution is not a suitable choice to describe the joint feature space. The scales of the two plots in Figure \ref{fig:feature_U01b} are also different. The variation of the features of correct respondents is much smaller than that of incorrect respondents. The main reason for this phenomenon is that there are more ways to solve the problem incorrectly than correctly. Item U01b requires the respondents to create a new folder and to move some emails to the new folder. Among the incorrect respondents, some moved emails but didn't create a new folder while some created a new folder but didn't move the emails correctly. There are also some respondents who didn't respond seriously---they took fewer than five actions before moving to the next item. As shown in the right panel of Figure \ref{fig:feature_U01b}, respondents with similar behaviors are located close to each other in the feature space.


\subsection{Reconstruction of Derived Variables}

We demonstrate in this subsection that the features extracted from the proposed procedure retain a substantial amount of information of the response processes. To be more specific, we show that various variables directly derived from the processes can be reconstructed by the extracted features.

We define a derived variable as a binary variable indicating whether an action or a combination of actions appears in the process. For example, whether the first dropdown menu is set to ``Photography'' is a derived variable of the item described in the introduction. The binary response outcome is also a derived variable since it is entirely determined by the response process.
In our data, 93 derived variables, including 14 item response outcomes, are considered. 

Similar to the simulation study, we examine how well the derived variables can be reconstructed through a prediction procedure. We use logistic regression to model the relation between a derived variable and the principal features of the corresponding item.
For each derived variable, 80\% of the respondents are randomly sampled to form the training set and the remaining 20\% form the test set. We fit the model on the training set and predict the derived variable for each respondent in the test set. Specifically, the derived variable is predicted as 1 if the fitted probability is greater than 0.5 and 0 otherwise. Prediction accuracy on the entire test set is calculated for evaluation.

As shown in Table \ref{table:derived_var}, for all the derived variables, the prediction accuracy is higher than 0.80. For 75 out of 93 variables, the accuracy is higher than 0.90. Thirty five variables are predicted nearly perfectly (prediction accuracy greater than $0.975$). These results manifest that the extracted features carry a significant amount of information in the action sequences. We demonstrate in the remaining subsections that the extracted features is useful for assessing respondents' competency and behaviors.


\begin{table}
\centering
\caption{Distribution of the out-of-sample prediction accuracy for 93 derived variables.}\label{table:derived_var}
\begin{tabular}{cccccc}
\hline
Accuracy & $(0.80, 0.85]$ & (0.85, 0.90] & (0.90, 0.95] & (0.95, 0.975] & (0.975, 1.00] \\
\hline
Counts & 5 & 13 & 28 & 12 & 35\\
\hline
\end{tabular}
\end{table}


\subsection{Variable Prediction Based on a Single Item}
Item responses (both outcome and process) of an item reflect respondents' latent traits, which affect their overall performance in a test.
Therefore, each item response should have some predicting power of the responses of other items and the overall competency. Process data contain more detailed information of respondents' behaviors than a single binary outcome. We expect that the prediction based on the response process is more accurate than that solely based on the final outcome. In this subsection, we assess information in the response processes of a single item via the prediction of the binary response outcomes of other items as well as the numeracy and the literacy scores. 

Given the final outcome and the response process of an item, say item $j$, we model their relation with the predicted variable by a generalized linear model
\begin{equation}\label{eq:pred_model}
g(\mu) = \bm \eta_j^\top \bm \beta,
\end{equation}
where $\mu$ is the expectation of the predicted variable, $g$ is the link function, $\bm \eta_j$ is a vector of covariates related to item $j$, which will be specified later, and $\bm \beta$ is the coefficient vector. If the predicted variable is the binary outcome of item $j'$, $g(\mu) = \log\left(\mu/(1-\mu)\right)$ is the logit link and $\mu$ is the probability of answering the item correctly. If the predicted variable is the literacy or numeracy score, $g$ is the identity link and \eqref{eq:pred_model} becomes linear regression.

Let $z_j$ denote the binary outcome and let $\bm\theta_j$ denote the features extracted from the response process of item $j$. We consider two choices of $\bm\eta_j$ for a given predicted variable, $\bm \eta_j = (1, z_j)^\top$ and $\bm \eta_j = (1, z_j, \bm\theta_j^\top, z_j \bm\theta_j^\top)^\top$. The first choice only uses the binary outcome for prediction. The second uses both the outcome and the response process. We call the model with these two choices of covariates the baseline model and the process model, respectively. It turns out that the information in the baseline model is very limited, especially when the correct rate of item $j$ is close to 0 or 1.

For a given predicted variable, two thirds of the available respondents are randomly sampled to form the training set. The remaining one third are evenly split to form the validation and the test set. Both the baseline model and the process model are fit on the training set. We add $L_2$ penalties on the coefficient vector $\bm \beta$ in the process model to avoid overfitting. The penalty parameter is chosen by examining the prediction performance of the resulting model on the validation set. Specifically, a process model is fitted for each candidate value of the penalty parameter. The one that produces the best prediction performance on the validation set is chosen to obtain the final process model for comparing with the baseline model. The evaluation criterion is prediction accuracy for outcome prediction and out-of-sample $R^2$ ($\text{OSR}^2$) for score prediction. $\text{OSR}^2$ is defined to be the square of the Pearson correlation between the predicted and true values. A higher $\text{OSR}^2$ indicates better prediction performance.

\subsubsection{Outcome Prediction Results}

\begin{figure}[htb]
\includegraphics[width=\textwidth]{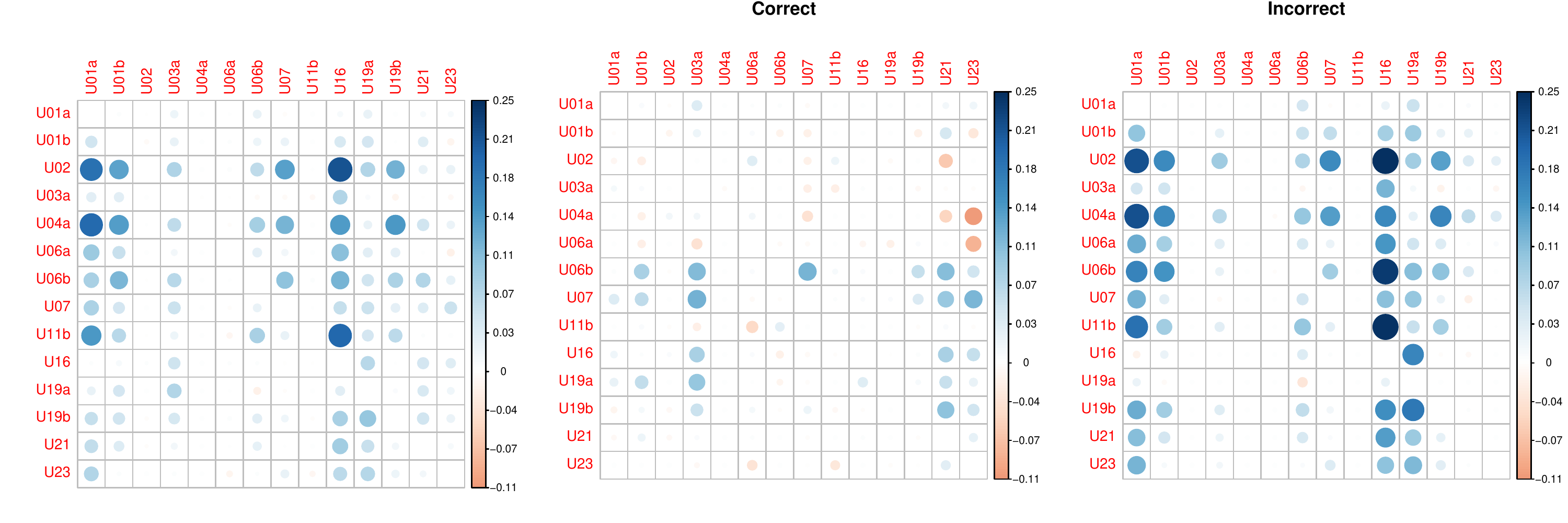}
\caption{Difference of the cross-item outcome prediction accuracy for the process model and the response model.}\label{fig:outcome_pred}
\end{figure}

Figure \ref{fig:outcome_pred} presents the results of outcome prediction. The plot in the left panel gives the improvement in the out-of-sample prediction accuracy of the process model over that of the baseline model for all item pairs. The entry in the $i$-th row and the $j$-th column gives the result for predicting item $j$ by item $i$. For many item pairs, adding the features extracted from process data improves the prediction. To further examine the improvements, for the task of predicting the outcome of item $j'$ by item $j$, we calculate the prediction accuracy separately for the respondents who answered item $j$ correctly and for those who answered incorrectly. The improvements for these two groups are plotted respectively in the middle and the right panels of Figure \ref{fig:outcome_pred}. The improvement is more significant for the incorrect group in both the number of item pairs that have improvement and the magnitude of the improvement. As we mentioned previously, the incorrect response processes are more diverse than the correct ones, thus providing more information about the respondents. Misunderstanding the item requirements and lack of basic computer skills often lead to an incorrect response. Carelessness and inattentiveness are also possible causes of an incorrect answer. These differences can be reflected in the extracted features as illustrated in Figure \ref{fig:feature_U01b}. Therefore, including these features in the model helps the prediction more for the incorrect group than for the correct group.

\subsubsection{Numeracy and Literacy Prediction Results}

\begin{figure}[htb]
\centering
\includegraphics[width=\textwidth]{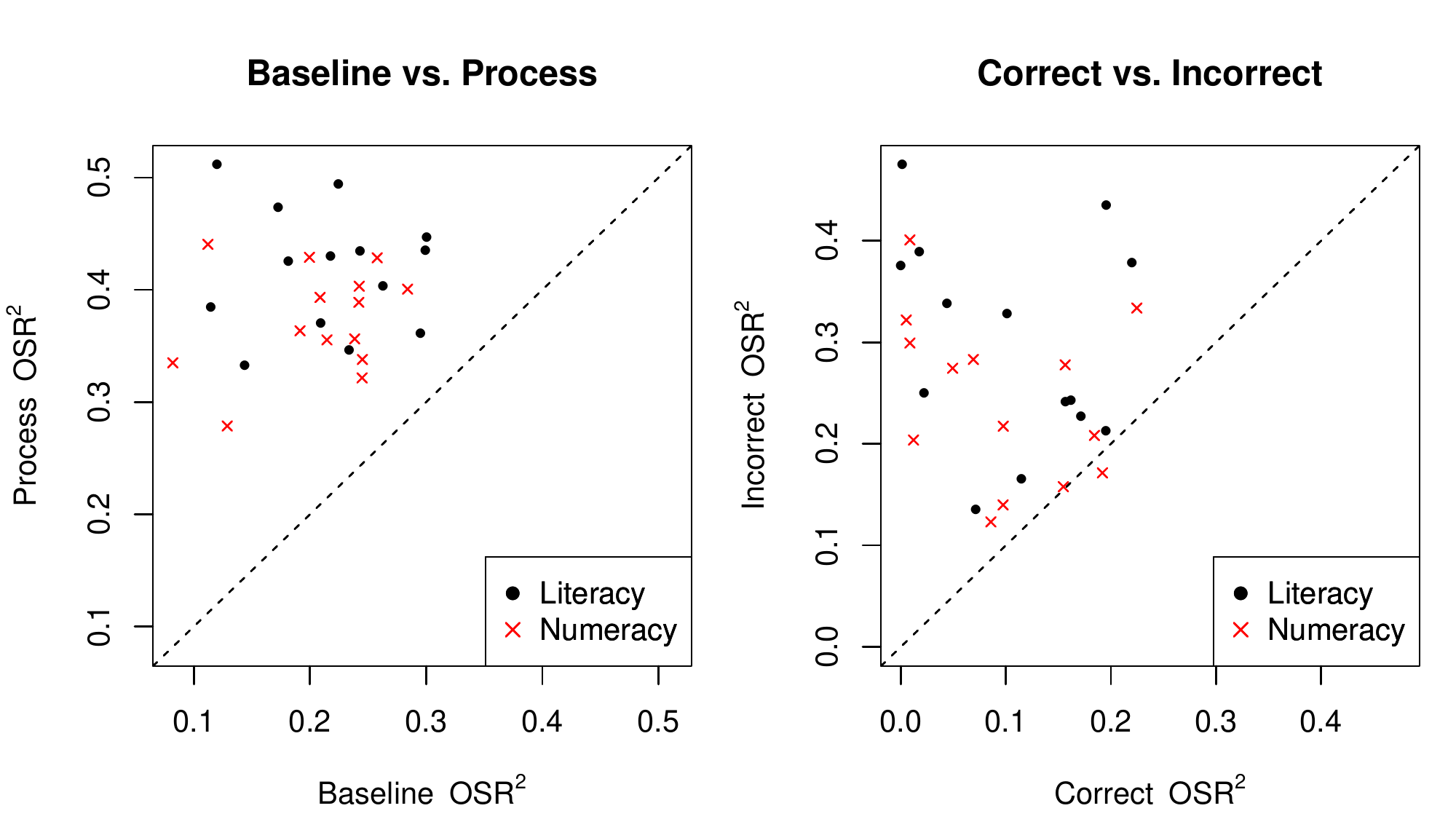}
\caption{Left: OSR${}^2$ of the baseline and process model on the test set. Right: OSR${}^2$ of the process model for the correct and incorrect groups.}\label{fig:score_pred}
\end{figure}

Numeracy and literacy score prediction results are displayed in Figure \ref{fig:score_pred}. In the left panel, we plot the $\text{OSR}^2$ of the process model against that of the baseline model. For both literacy and numeracy, regardless of the item used for prediction, the process model produces a higher $\text{OSR}^2$ than the baseline model.
Although the PSTRE items are not designed for measuring these two competencies, the response processes are helpful for predicting the scores. 
To further examine the results, for each item-score pair, we again group the respondents according to their item response outcome and calculate the $\text{OSR}^2$ of the process model for the two groups separately. The $\text{OSR}^2$ for the incorrect group is plotted against that for the correct group in the right panel of Figure \ref{fig:score_pred}. Similar to the outcome prediction, the prediction performance for the incorrect group is usually much better than that for the correct group since action sequences corresponding to incorrect answers are often more diverse and informative than those corresponding to correct answers.

%


\subsection{Prediction Based on Multiple Items}
In this subsection, we examine how the improvement in prediction performance brought by process data aggregates as more items are incorporated in the prediction. The variables of interest are age, gender, and literacy and numeracy scores.

We only consider the 3,645 respondents who responded to all 14 PSTRE items in this experiment. The respondents are randomly split into training, validation, and test sets. The sizes of the three sets are 2645, 500, and 500, respectively. The split is fixed for estimating and evaluating all models in this experiment. 

We still consider model \eqref{eq:pred_model} for prediction. The logit link, i.e., logistic regression, is used for gender prediction and linear regression for other variables. In this experiment, the covariate vector $\bm \eta$ incorporates information from multiple items. 
Given a predicted variable and a set of available items, a baseline model and a process model are considered for each variable. 
For the baseline model, $\bm \eta$ consists of only the final outcomes, while for the process model it also includes the first 20 principal features for each available item. 
Let $S_m = \{j_1, \ldots, j_m\}$ denote the set of the indices of available items. The predictor for the baseline model is $\bm \eta = (1, z_{j_1}, \ldots, z_{j_m})^\top$ and that of the process model is $\bm \eta = (1, z_{j_1}, \ldots, z_{j_m}, \bm \theta_{j_1}, \ldots, \bm \theta_{j_m})^\top$ where $\bm \theta_{j} \in \mathbb{R}^{20}$ is the first 20 principal features for item $j$. 
We start from an empty item set and add one item to the set at a time. That is, for a given predicted variable, a sequence of 14 baseline models and 14 process models are fitted. The order of items being added to the model is determined by forward Akaike information criterion (AIC) selection for the 14 outcomes on the training set. Specifically, for a given $m$, $S_m$ contains the items whose outcomes are the first $m$ variables selected by the forward AIC selection among all 14 outcomes. 
We use prediction accuracy as the evaluation criterion for gender prediction and $\text{OSR}^2$ for other variables.

\subsubsection{Numeracy and Literacy Prediction Results}

\begin{figure}
\centering
\includegraphics[width=\textwidth]{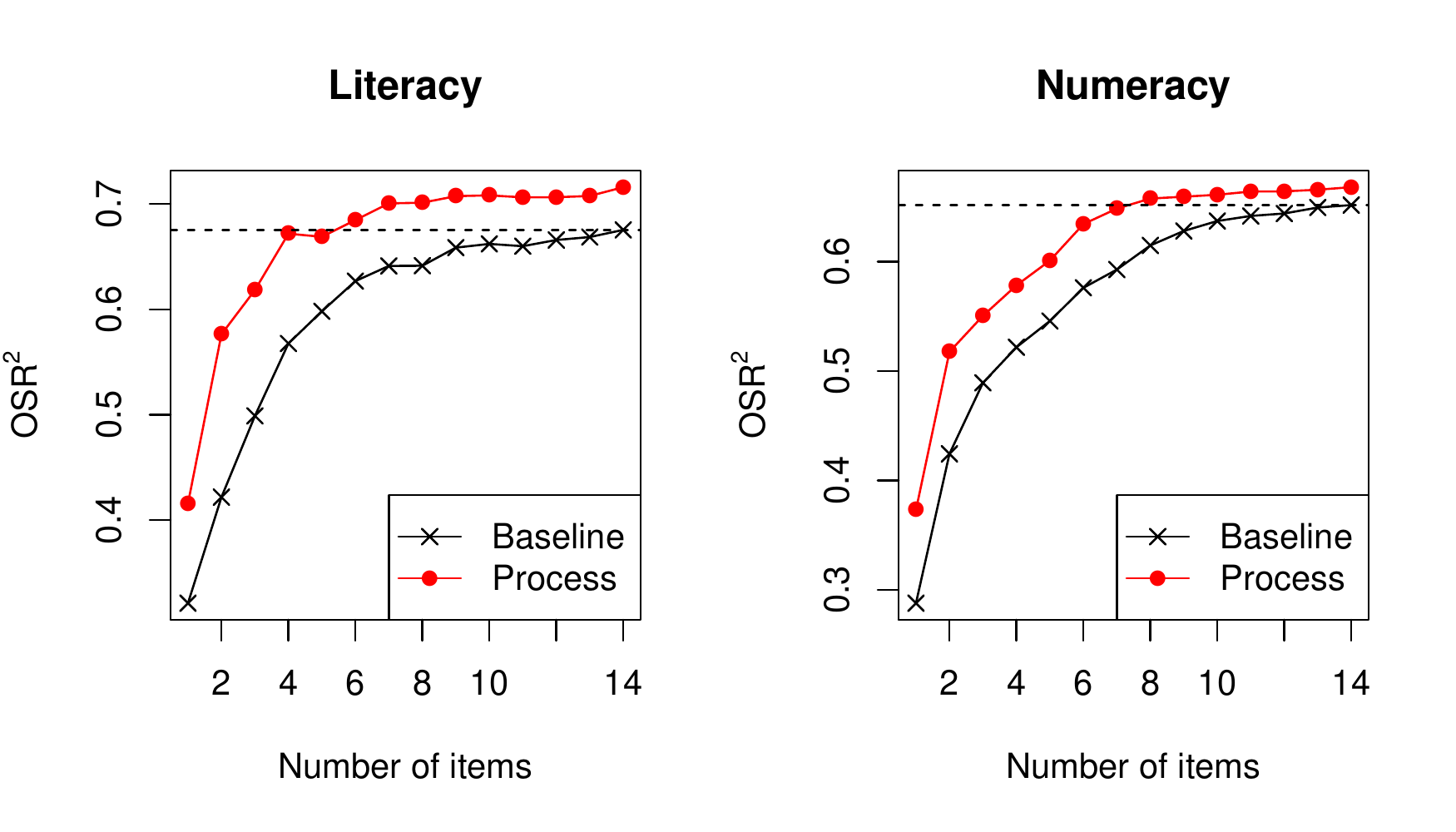}
\caption{OSR${}^{2}$ of the baseline and process model with various number of items.}\label{fig:score_pred_all}
\end{figure}

In Figure \ref{fig:score_pred_all}, $\text{OSR}^2$ for predicting literacy and numeracy scores is plotted against the number of available items. For both the process model and the baseline model, the prediction of the numeracy and the literacy improves as responses from more items are available. Regardless of the number of available items, the process model outperforms the baseline model in both literacy and numeracy score predictions, although the difference becomes smaller as the the number of available items increases.
Notice that the $\text{OSR}^2$ of the process model based on only two items roughly equals the $\text{OSR}^2$ of the baseline model based on four items.
These results imply that properly incorporating process data in data analysis can exploit the information in items more efficiently and that the incorporation is especially beneficial when a small number of items are available.

The PSTRE item responses have some predicting power of literacy and numeracy. This is not surprising as literacy and numeracy are related to the understanding of the PSTRE item description and material. 
In our case study, PSTRE items are more related to literacy than numeracy---the $\text{OSR}^2$ of literacy score models is usually higher than that of the corresponding numeracy score model. The number of items needed in the process model to achieve a similar $\text{OSR}^2$ obtained in the baseline model with all 14 items is five for literacy and eight for numeracy.



\subsubsection{Background Variable Prediction Results}
\begin{figure}
\centering
\includegraphics[width=\textwidth]{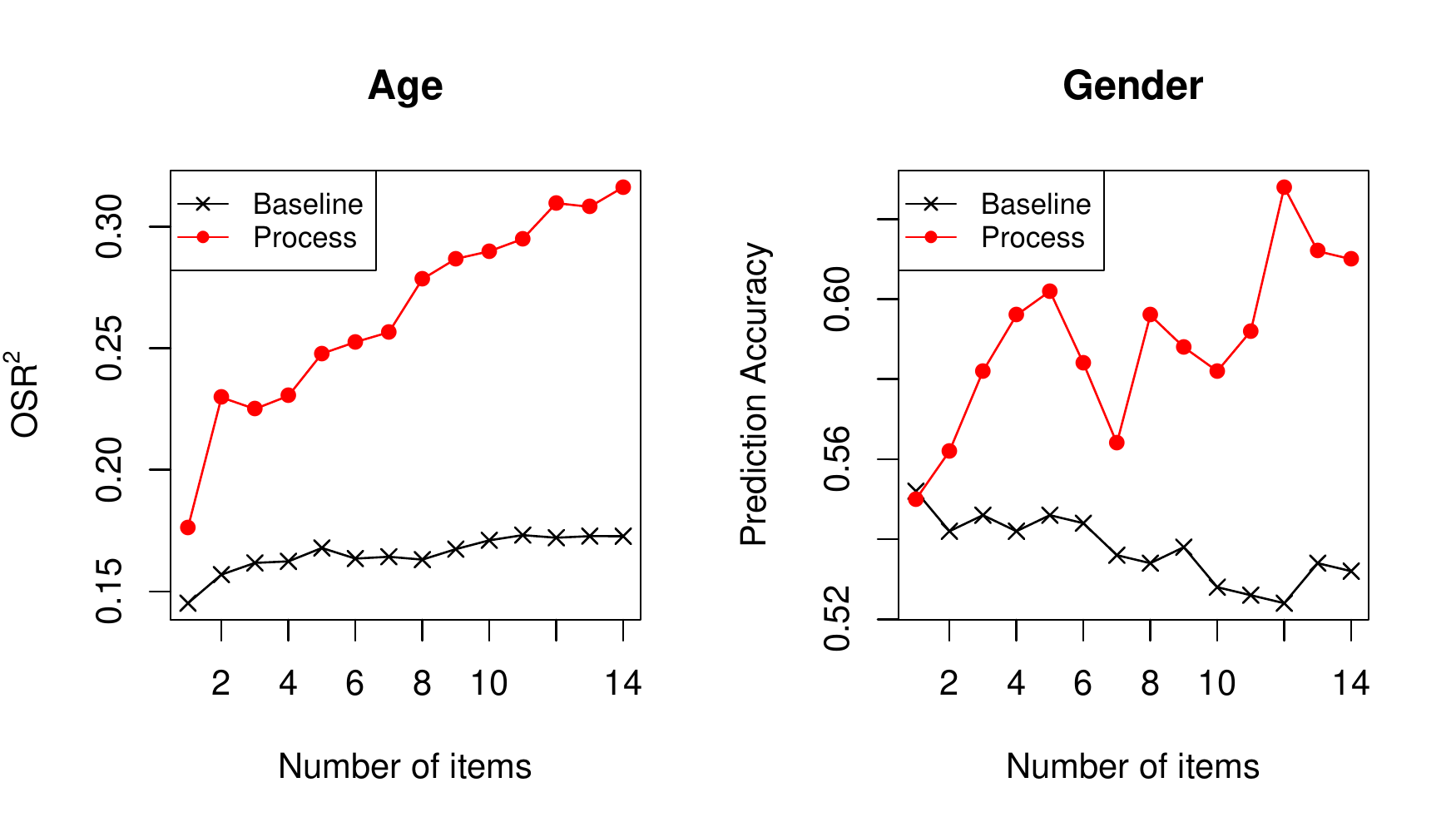}
\caption{Prediction results for age and gender.}\label{fig:bg_pred_all}
\end{figure}

\begin{figure}
\centering
\includegraphics[width=0.45\textwidth]{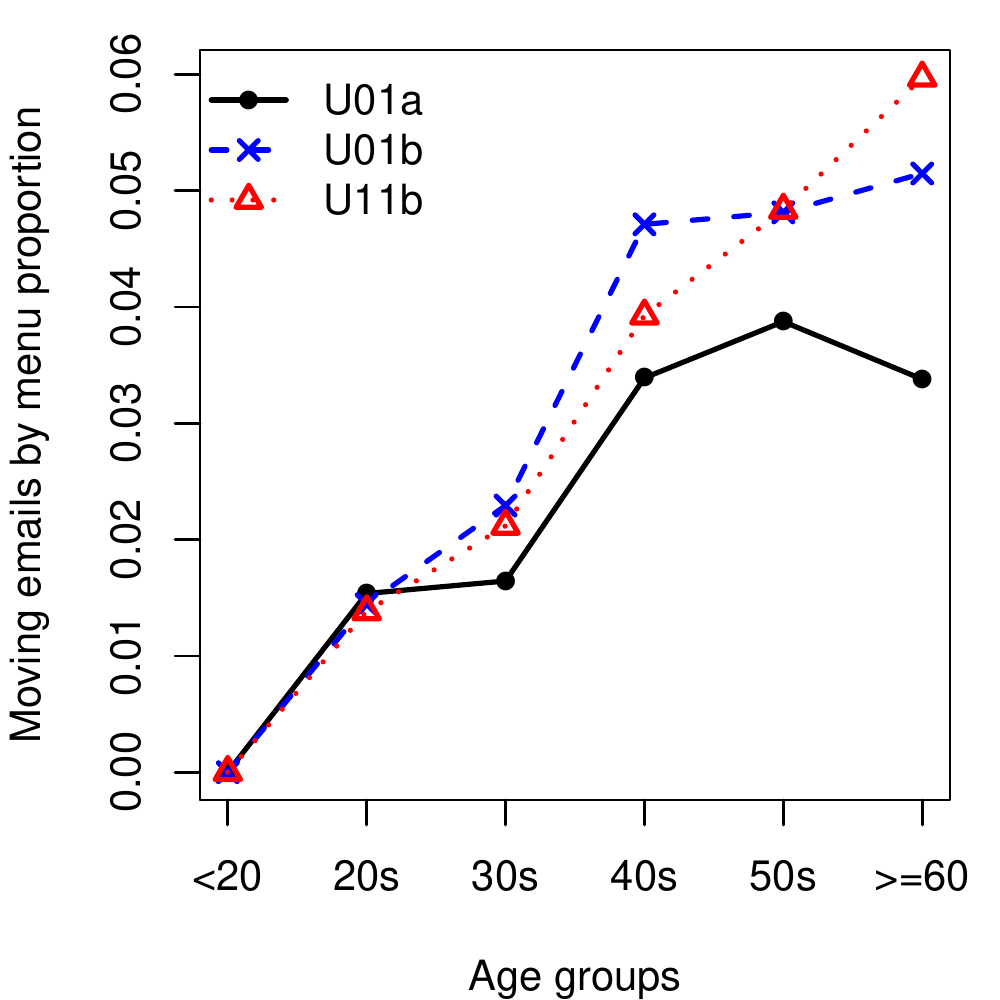}
\includegraphics[width=0.45\textwidth]{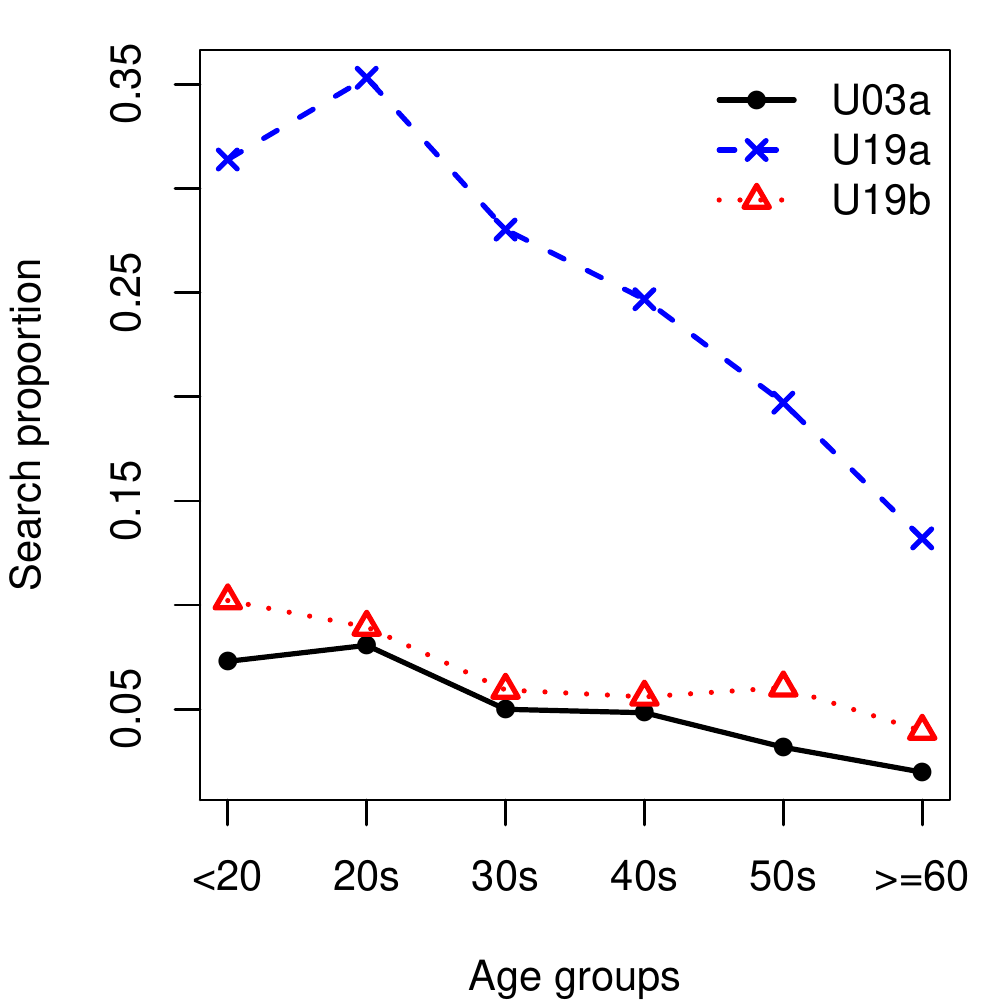}
\caption{Left: Proportion of respondents moving emails by menu in different age groups. Right: Proportion of respondents using ``Search'' in different age groups.}\label{fig:age_evidence}
\end{figure}

Figure \ref{fig:bg_pred_all} presents the results for predicting age and gender. Adding more items in the baseline model barely improves the $\text{OSR}^2$ for predicting age while in the process model the quantity increases as more items are included and it is about twice as high as that of the baseline model when all 14 items are included. These results show that respondents at different age behave differently in solving PSTRE items and that response processes can reveal the differences significantly better than final outcomes. A closer examination of the action sequences shows that younger respondents are more likely to use drag and drop actions to move emails while older respondents tend to move emails by using email menu (left panel of Figure \ref{fig:age_evidence}). Also, older respondents are less likely to use ``Search'' in spreadsheet environment (right panel of Figure \ref{fig:age_evidence}).

As for gender, the highest prediction accuracy of the baseline models is 0.55, which is only 0.02 higher than the proportion of female respondents in the test set. The prediction accuracy of the process model is almost always higher than that of the corresponding baseline model and it can be as high as 0.63. These observations imply that female and male respondents have similar performance in PSTRE items in terms of final outcomes, but there are some differences in their response processes. In our data, male respondents are more likely to use sorting tools in spreadsheet environment as shown in Table \ref{table:gender_sort}. The p-value for the $\chi^2$ test of independence between gender and whether ``Sort'' is used is less than $10^{-6}$ for the three items with spreadsheet environment.

\begin{table}
\centering
\caption{Contingency tables of gender and whether ``Sort'' is used in U03a, U19a, and U19b.}\label{table:gender_sort}
\begin{tabular}{c cc c cc c cc}
\hline
\multirow{2}{*}{Gender} & \multicolumn{2}{c}{U03a} &  & \multicolumn{2}{c}{U19a} & & \multicolumn{2}{c}{U19b}\\
\cline{2-3}\cline{5-6}\cline{8-9}
 & Yes & No & & Yes & No & & Yes & No\\
\hline
Male & 418 & 1238 & & 365 & 1291 & & 661 & 995\\
Female & 359 & 1630 & & 311 & 1678 & & 564 & 1425\\
\hline
\end{tabular}
\end{table}

\section{Concluding Remarks}\label{sec:discussion}
In this article, we presented a method to extract latent features from response processes. The key step of the method is to build an action sequence autoencoder for a set of response processes. We showed through a case study of the process data of PSTRE items in PIAAC 2012 that the extracted features improve the prediction of response outcomes, and literacy and numeracy scores. 

It is possible to build neural networks that predict a response variable directly from an action sequence. These neural networks are often the combination of an RNN and a feed-forward neural network. In this way, we possibly need to fit separate models for each response variable and each of the models involves RNNs. Fitting models with RNN components is generally computationally expensive because of its recurrent structure. With the feature extraction method, we only need to fit a single model (the action sequence autoencoder) that involves RNNs, and then fit a (generalized) linear model or a feed-forward neural network for each variable of interest. The prediction performance of the two approaches are often comparable. The approach without feature extraction may perform worse than the approach with feature extraction due to overfitting.

Computer log files of interactive items often include time stamps of actions. The time elapsed between two consecutive actions may also provide extra information about respondents and can be useful in educational and cognitive assessments. The current action sequence autoencoder does not make use of this information. Further study on incorporating time information in the analysis of process data is a potential future direction.

\section{Acknowledgment}
The authors would like to thank Educational Testing Service and Qiwei He for providing the data, and Hok Kan Ling for cleaning it.

\bibliographystyle{apacite}
\bibliography{seq2seq}

\appendix
\section{Structures of the LSTM Unit and the GRU}
\subsection{LSTM Unit}
Using the notation in Section \ref{sec:rnn}, the LSTM unit computes the hidden states and outputs in time step $t$ as follows
\begin{gather*}\label{eq:lstm}
\bm z_t = \sigma (\bm q_1 + \bm W_1 \bm x_t + \bm U_1 \bm m_{t-1}),\\
\bm r_t = \sigma (\bm q_2 + \bm W_2 \bm x_t + \bm U_2 \bm m_{t-1}), \\
\tilde{\bm c}_t = \tanh (\bm q_3 + \bm W_3 \bm x_t + \bm U_3 \bm m_{t-1}), \\
\bm c_t = \bm z_t \star \bm c_{t-1} + \bm r_t \star \tilde{\bm c_t}, \\
\bm v_t = \sigma (\bm q_4 + \bm W_4 \bm x_t + \bm r_t \star \bm U_4 \bm m_{t-1}), \\
\bm m_t = \bm v_t \star \tanh (\bm c_t),\\
\bm y_t = \bm m_t,
\end{gather*}
where $\star$ denotes element-wise multiplication, $\bm z_t$, $\bm r_t$, $\bm v_t$, and $\bm c_t$ are called the forget gate, input gate, output gate, and cell state of an LSTM unit, respectively, and $\bm q_i, \bm W_i, \bm U_i, i = 1,2,3,4$, are parameters. Both $\sigma (x) = 1/\{1+ \exp(-x)\}$ and $\tanh (x) = \{\exp(x) - \exp(-x) \}/\{\exp(x) + \exp(-x)\}$ are element-wise activation functions.

\subsection{GRU}
Using the notation in Section \ref{sec:rnn}, the GRU computes the hidden states and outputs in time step $t$ as follows
\begin{gather*}\label{eq:lstm}
\bm z_t = \sigma (\bm q_1 + \bm W_1 \bm x_t + \bm U_1 \bm m_{t-1}),\\
\bm r_t = \sigma (\bm q_2 + \bm W_2 \bm x_t + \bm U_2 \bm m_{t-1}), \\
\tilde{\bm m}_t = \tanh (\bm q_3 + \bm W_3 \bm x_t + \bm U_3 (\bm r_t \star \bm m_{t-1})), \\
\bm m_t = (1 - \bm z_t) \star \bm m_{t-1} + \bm z_t \star \tilde{\bm m}_t, \\
\bm y_t = \bm m_t,
\end{gather*}
where $\star$ denotes element-wise multiplication, $\bm z_t$ and $\bm r_t$ are called the update gate and reset gate of a GRU, respectively, and $\bm q_i, \bm W_i, \bm U_i, i = 1,2,3$, are parameters. Both $\sigma (x) = 1/\{1+ \exp(-x)\}$ and $\tanh (x) = \{\exp(x) - \exp(-x) \}/\{\exp(x) + \exp(-x)\}$ are element-wise activation functions.

\end{document}